\documentclass[lettersize,journal]{IEEEtran}
\usepackage{amsmath,amsfonts}
\usepackage{algorithmic}
\usepackage[ruled,linesnumbered]{algorithm2e}
\usepackage{array}
\usepackage{bbm}
\usepackage{bm}
\usepackage[caption=false,font=normalsize,labelfont=sf,textfont=sf]{subfig}
\usepackage{textcomp}
\usepackage{stfloats}
\usepackage{url}
\usepackage{verbatim}
\usepackage{graphicx}
\usepackage{cite}
\usepackage{tabularx,booktabs}
\usepackage{multirow}
\usepackage{multicol}
\usepackage{colortbl} 
\usepackage{xcolor}
\usepackage{hyperref}
\usepackage{amssymb}
\usepackage{amsthm}
\usepackage{mathrsfs}
\usepackage{cite}
\usepackage{bbding}

\definecolor{lightblue}{RGB}{204,186,230}
\definecolor{citeblue}{RGB}{32,67,192}
\newcommand{\tx}{\textcolor{black}}
\newcommand{\ttx}{\textcolor{black}}

\hypersetup{
	colorlinks=true,
	linkcolor=red,
	anchorcolor=blue,
	citecolor=green}

\newcommand{\bb}{\boldsymbol}

\newcommand{\e}{et al.}

\newcommand{\R}{\mathbb{R}}
\newcommand{\cS}{\mathcal{S}} 
\newcommand{\cA}{\mathcal{A}}

\renewcommand{\epsilon}{\varepsilon}
\renewcommand{\phi}{\varphi}

\newtheorem{theorem}{ \bf Theorem}
\newtheorem{lemma}{\bf Lemma}
\newtheorem{assumption}{\bf Assumption}
 
\newtheorem{proposition}{\bf Proposition}
\newtheorem{definition}{\bf Definition}

\newcolumntype{L}[1]{>{\raggedright\arraybackslash}p{#1}}
\newcolumntype{C}[1]{>{\centering\arraybackslash}p{#1}}
\newcolumntype{R}[1]{>{\raggedleft\arraybackslash}p{#1}}

\makeatletter
\newcommand{\removelatexerror}{\let\@latex@error\@gobble}
\renewcommand{\maketag@@@}[1]{\hbox{\m@th\normalsize\normalfont#1}}%
\makeatother
\usepackage{makeidx}
\makeindex

\begin{document}
\title{ LEASE: Offline Preference-based Reinforcement Learning with High Sample Efficiency}
	\author{
    Xiao-Yin Liu,~\IEEEmembership{} Guotao Li*,~\IEEEmembership{} Xiao-Hu Zhou,~\IEEEmembership{}
 Zeng-Guang Hou*,~\IEEEmembership{Fellow, IEEE}
		\thanks{This work is funded by the National Natural Science Foundation of China under (Grant 62473365, Grant U22A2056, and Grant 62373013), and the Beijing Natural Science Foundation under (Grant L232021 and L242101). (*Corresponding authors: Guotao Li and Zeng-Guang Hou).}
		\thanks{Xiao-Yin Liu, Guotao Li and Xiao-Hu Zhou are with the State Key Laboratory of Multimodal Artificial Intelligence Systems, Institute of Automation, Chinese Academy of Sciences, Beijing 100190, China, and also with the School of Artificial Intelligence, University of Chinese Academy of Sciences, Beijing 100049, China. (e-mail: liuxiaoyin2023@ia.ac.cn, guotao.li@ia.ac.cn, xiaohu.zhou@ia.ac.cn).}
  
		\thanks{Zeng-Guang Hou is with the State Key Laboratory of Multimodal Artificial Intelligence Systems, Institute of Automation, Chinese Academy of Sciences, Beijing 100190, China, also with the School of Artificial Intelligence, University of Chinese Academy of Sciences, Beijing 100049, China, and also with CASIA-MUST Joint Laboratory of Intelligence Science and Technology, Institute of Systems Engineering, Macau University of Science and Technology, Macao, China. (e-mail: zengguang.hou@ia.ac.cn).}
	}
	\markboth{}%
	{Shell \MakeLowercase{\textit{et al.}}: A Sample Article Using IEEEtran.cls for IEEE Journals}
	\maketitle
	
	\begin{abstract}
Offline preference-based reinforcement learning (PbRL) offers an effective approach to addressing the challenges of designing rewards and mitigating the high costs associated with online interaction. However, since labeling preference needs real-time human feedback, acquiring sufficient preference labels is challenging. To solve this, this paper proposes an off\textbf{L}ine pr\textbf{E}ference-b\textbf{A}sed RL with high \textbf{S}ample \textbf{E}fficiency (\texttt{LEASE}) algorithm, where a learned transition model is leveraged to generate unlabeled preference data. Considering the pretrained reward model may generate incorrect labels for unlabeled data, we design an uncertainty-aware mechanism to ensure the performance of the reward model, where only high-confidence and low-variance data are selected. 
Moreover, the generalization bound of the reward model is provided to analyze the factors influencing reward accuracy, and the policy learned by \texttt{LEASE} has a theoretical improvement guarantee. The above developed theory is based on a state-action pair, which can be easily combined with other offline algorithms.
The experimental results show that \texttt{LEASE} can achieve comparable performance to the baseline under fewer preference data without online interaction. 
	\end{abstract}


	\section{Introduction}\label{sec:introduction}
\IEEEPARstart{R}{einforcement} learning (RL) has been widely applied in robot control \cite{haarnoja2024learning,radosavovic2024real} and healthcare \cite{li2021toward,wu2022human} fields. However, in the real-world scenarios, RL faces two serious challenges: 1) Since RL learns the optimal policy through trial and error, it is dangerous for agent to interact directly with environment, especially human-in-loop control \cite{levine2020offline,fan2025coevolutionary}; 2) Due to the complexity of human cognitive and control system, it is difficult to design a quantitative reward function to accurately reflect the human intention or preference \cite{hadfield2017inverse}. Much research has been conducted to solve the above two challenges.

Offline RL, learning policy from previously collected dataset without interaction with the environment, has become an effective way for \textit{challenge 1} \cite{kumar2020conservative,kostrikov2021offline}. 
However, due to limited data coverage, offline RL suffers from the distribution shift between the learned policy and the behavior policy \cite{kumar2019stabilizing,10237048}, which leads to poor agent performance. To address this problem, offline RL either directly constrains the policy close to the behavior policy \cite{fujimoto2019off,10516602} or penalizes the value function for out-of-distribution (OOD) data \cite{kumar2020conservative,kostrikov2021offline,11078293}. The above methods improve the performance of the agent by incorporating conservatism into policy learning.

Preference-based RL (PbRL), also known as RL from human feedback (RLHF), which learns the reward function from human preference data without the demand for tedious hand-engineered reward design \cite{christiano2017deep,9448304}, has been developed recently for \textit{challenge 2}. However, compared with data $(s,a,s')$, the collecting cost of preference data is higher since it demands human real-time feedback to label preference, which often brings high sample complexity (low sample efficiency) \cite{liang2022reward,park2022surf}. Some data augmentation techniques, such as generating unlabeled data through interaction with a simulator \cite{park2022surf} or temporal cropping \cite{hu2024query}, have been proposed to solve the problem of the limitation of preference data. However, the above methods need online interaction with the environment (simulator) and still demand large human feedback. Thus, in PbRL, the sample efficiency is still of utmost importance.

In offline PbRL settings, existing methods \cite{shin2021offline,kim2023preference,zhang2024flow} typically involve two steps: reward learning and policy learning. They learns the policy based on an offline static dataset without a reward and preference dataset, and introduce transformer \cite{kim2023preference} or diffusion model \cite{zhang2024flow} to improve the performance of the reward model and agent performance. However, the theoretical guarantees of reward generalization and policy improvement are not provided. Therefore, this paper aims to design a high sample efficiency offline PbRL algorithm that can achieve comparable performance with the baseline from fewer preference data without interaction with the environment, and develop the related theory for reward generalization bound and policy improvement guarantee. The key questions can be summarized as follows:
\begin{enumerate}[]
    \item In the algorithm, the collected preference data fails to cover all state-action regions, and the collection cost is too high. How to use the limited preference data to learn an accurate reward model and guarantee agent performance?
   \item  In theory, what factors affect the reward generalization, and what is the detailed relationship between the offline RL algorithm itself, the reward gap, and the improvement of policy performance?
\end{enumerate}

For question 1, motivated by model-based offline RL, which learns the environment model to broaden data coverage \cite{yu2020mopo,liu2023micro}, we also train the transition model to achieve data augmentation (improve sample efficiency). Specifically, the agent generates two different trajectories through interaction with the learned transition model, and then the pretrained reward model generates pseudo-labels for these two trajectories. However, the preference model may generate incorrect pseudo-labels, which may lead to unstable training and low performance. For question 2, there are very few algorithms for offline PbRL theory. The most relevant work is \cite{zhan2024provable}, where the theory of policy improvement guarantee is established. However, the generalization bound for the reward model is not considered, and this theory is based on the whole trajectory rather than the state-action pair in offline algorithms. This is not conducive to the theoretical analysis of specific offline PbRL algorithms, since many offline RL algorithms are based on the state-action pair. 

\begin{figure*}
	\centering
	\includegraphics[width=0.9\textwidth]{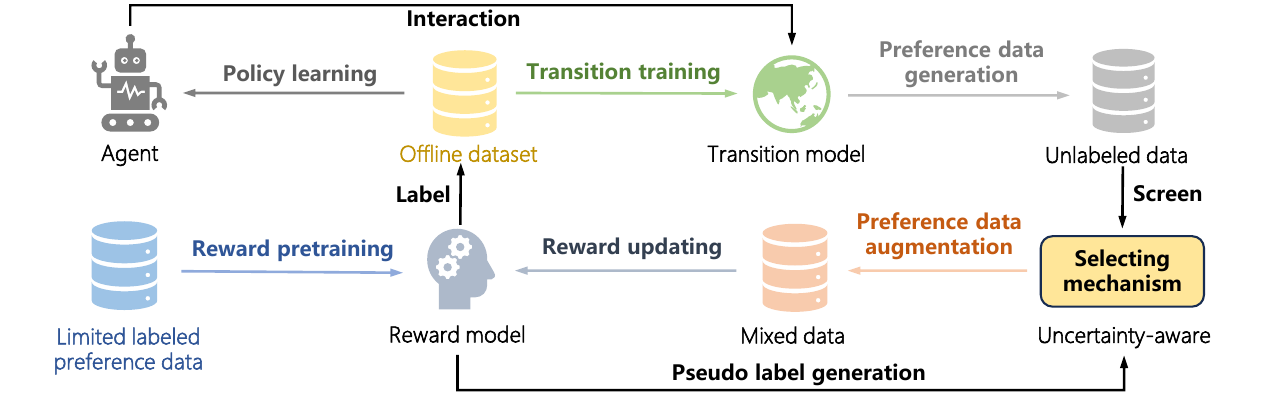}
	\caption{The learning framework of \texttt{LEASE}. The offline dataset without reward is used to train the transition model, and a limited labeled preference dataset is used to pretrain the reward model. Then, the generated unlabeled preference data is screened through the uncertainty-aware selection mechanism. The reward model is updated based on the labeled and generated dataset. Finally, the agent learns the policy based on an offline dataset and a learned reward model.}. 
	\label{fig1}
\end{figure*}

To solve the above problems, this paper proposes a novel off\textbf{L}ine pr\textbf{E}ference-b\textbf{A}sed RL with a high \textbf{S}ample \textbf{E}fficiency algorithm (\texttt{LEASE}), where a selecting mechanism is designed to guarantee the quantity of the generated dataset. This selection mechanism is based on the principles of high confidence and low uncertainty, and can effectively reduce the error of pseudo-labels. Moreover, we develop a new generalization bound for the reward model and provide a theory of policy improvement guarantee for offline PbRL based on the state-action pair. Fig. \ref{fig1} shows the learning framework of \texttt{LEASE}. The theoretical and experimental results show that \texttt{LEASE} has a policy improvement guarantee and can achieve comparable performance to the baseline under fewer preference data on the D4RL benchmark. The contributions are given below: 
\begin{enumerate}[]
    \item A novel learning framework, named \texttt{LEASE}, is proposed for offline PbRL, where an innovative preference augmentation technique is utilized to improve sample efficiency.
   \item An uncertainty-aware mechanism is designed for screening the generated preference data to guarantee the stability of reward training and improve the accuracy of reward prediction.
   \item The generalization bound of the reward model and the theory of policy improvement based on the state-action pair are developed to analyze factors that influence reward and policy performance.
\end{enumerate}

The framework of this paper is as follows: Section \ref{sec:Related work} introduces the related works about offline PbRL. Section \ref{Preliminaries} provides the basic notations for the algorithm and illustrates the two main problems that this paper aims to solve. Section \ref{reward} introduces the details of the designed selection mechanism and derives the generalization bound for the reward model. Section \ref{policy} illustrates the implementation details of the proposed method and provides theory for offline PbRL. Section \ref{experiments} conducts experiments to validate the effectiveness of \texttt{LEASE} and analyze the effects of each part on agent performance. Section \ref{discussion} further discusses the offline PbRL theory and algorithm. Section \ref{conclusion} summarizes the entire work.

	\section{Related Works}\label{sec:Related work}
\ttx{This section provides a brief overview of the algorithms and related theories in offline PbRL, and discusses the differences and advantages between our work and previous associated works.}

\textbf{The algorithm for offline PbRL.} Offline PbRL eliminates the demand for interaction with the environment and handcrafted-designed rewards. 
OPAL \cite{shin2021offline} is the first algorithm that combines offline RL and PbRL. PT \cite{kim2023preference} utilized a transformer-based architecture to design a preference model capable of generating non-Markovian rewards. OPPO \cite{kang2023beyond} directly optimized the policy in a high-level embedding space without learning a separate reward function. However, the above methods demand a large amount of preference data. FTB \cite{zhang2024flow} trains a diffusion model to achieve data augmentation, but this consumes a larger training cost. Instead, our method \texttt{LEASE} trains a simple transition model to generate unlabeled data and designs a selection mechanism to ensure the quality of the generated data, which can achieve superior performance under fewer datasets and less time.

\ttx{The aforementioned approach first learns an explicit reward function from preference data and then utilizes it to guide policy learning. However, in complex tasks, the reward function may contain substantial errors, which can degrade the performance of policy learning. Recently, preference-based direct policy learning has gained significant research attention \cite{rafailov2023direct,meng2024simpo,lanchantin2025diverse,ethayarajh2024kto}, as it directly optimizes the policy based on user preferences. The most representative work in this area is the Direct Preference Optimization (DPO) algorithm \cite{rafailov2023direct}, which formulates an initial preference optimization objective. Subsequent research has focused on designing different forms of preference optimization functions to further improve policy learning performance, including simpler preference optimization formulations \cite {meng2024simpo,lanchantin2025diverse} and optimization objectives based on prospect theory \cite{ethayarajh2024kto}. This direction constitutes a key focus of preference-based policy optimization.}

\textbf{The theory for offline PbRL.} There are a few algorithms that provide theoretical guarantees for offline PbRL, including the generalization bound of the reward model and the guarantee of policy improvement. Zhu \e \cite{zhu2023principled} studied offline PbRL, but the analysis is restricted to the linear model. Zhan \e \cite{zhan2024provable} extended to general function approximation, but the generalization analysis of the reward model is not provided, and the theory is based on the trajectory. \texttt{LEASE} gives the theoretical analysis for the reward model, and provides the theoretical guarantee for the policy based on the state-action pair. The generalization bound of the reward model has generality. It can be applied to methods that train a reward model using pseudo-labels, and the theory of policy improvement guarantees in offline PbRL can be easily combined with other offline RL approaches.

\textbf{Data selection techniques.} The data screening mechanism essentially belongs to the field of semi-supervised learning. The goal of semi-supervised learning is to use unlabeled data to improve model performance when labeled data is limited. Data selection techniques are used to filter the data with clean labels from a noisy dataset. Han \e \cite{han2018co}  selected unlabeled data with small losses from one network. Li \e \cite{lidividemix} used a two-component GMM to separate the dataset into a clean set and a noisy set. Rizve \e \cite{rizve2021defense} modeled the prediction uncertainty of unlabeled data to screen data. Xiao \e \cite{xiao2023promix} selected data with high confidence as clean data. Motivated by \cite{rizve2021defense} and \cite{xiao2023promix}, we trained an ensemble reward model to select data with high confidence and low variance to guarantee the quality of unlabeled data.

	\section{Preliminaries and Problem Formulation}\label{Preliminaries}
In this section, we provide the basic notations for reinforcement learning algorithms, including offline RL, preference-based RL, and model-based RL. Then, we introduce the two main problems that this paper aims to solve from both algorithmic and theoretical perspectives.

\subsection{Preliminaries for RL}
{\textbf{Offline reinforcement learning.}} The framework of RL is based on the Markov Decision Process (MDP) that is described by the tuple $\mathcal{M}=(\mathcal{S}, \mathcal{A}, R, T, \rho, \gamma)$, where $\cS$ is the state space, $\cA$ is the action space, $R: \cS\times \cA\rightarrow \R$ is the reward function, $T: \cS\times \cA\rightarrow \Delta({\cS})$ is the transition dynamics, $\rho$ is the initial state distribution, and $\gamma \in(0,1)$ is the discount factor \cite{levine2020offline}. The term $\Delta({\Omega})$ denotes the set of probability distributions over space $\Omega$. The goal of RL is to optimize the policy $\pi$ that maximizes the expected discounted return $J(\pi,R):=\mathbb{E}_{(s,a)\sim d_{T}^{\pi}(s,a)}[R(s,a)]/(1-\gamma)$, where $d_{T}^{\pi}(s, a):=d_{T}^{\pi}(s) \pi(a|s)$ is the state-action marginal distribution under the learned policy $\pi$. The discounted state marginal distribution $d_{T}^{\pi}(s)$ is denoted as $(1-\gamma)\sum_{t=0}^{\infty}\gamma^t\mathcal{P}(s_t = s|\pi)$, where $\mathcal{P}(s_t = s|\pi)$ is the probability of reaching state $s$ at time $t$ by rolling out $\pi$. The policy $\pi$ can be derived from $Q$-learning \cite{sutton1998introduction}, which learns the $Q$-function that satisfies Bellman operator: $\mathcal{T} Q(s,a):=R(s, a)+\gamma \mathbb{E}_{s^{\prime} \sim T_{\mathcal{M}}(s^{\prime} \mid s, a)}[\max_{a'\in \cA} Q(s', a')]$. 



The goal of offline RL is to learn a policy from offline dataset $\mathcal{D}_{\text{offline}}=\{\left(s_{i}, a_{i}, r_{i}, s_{i}^{\prime}\right)\}_{i=1}^{C}$ collected by the behavior policy $\mu$. The policy learning includes two parts: policy evaluation (minimizing the Bellman error) and policy improvement (maximizing the $Q$-function), that is
\begin{equation}\label{eq_1}
	\begin{aligned}
		&\widehat{Q} \leftarrow \arg \min _{Q} \mathbb{E}_{s, a, s^{\prime} \sim \mathcal{D}}\big[Q(s, a)-\widehat{\mathcal{T}} \widehat{Q}(s, a)\big]^{2},\\
		&\widehat{\pi} \leftarrow \arg \max _{\pi} \mathbb{E}_{s \sim \mathcal{D}, a \sim \pi}\big[\widehat{Q}(s, a)\big],
	\end{aligned}
\end{equation}
where $\mathcal{D}$ can be a fixed offline dataset or replay buffer generated by the current policy $\widehat{\pi}$ interacting with the environment. The operator $\widehat{\mathcal{T}}$ is the Bellman operator based on sample, denoted as $\widehat{\mathcal{T}}Q(s,a):=R(s, a)+\gamma \max_{a'\in\mathcal{A}}Q\left(s^{\prime}, a^{\prime}\right)$ \cite{kumar2020conservative}.



{\textbf{Preference-based reinforcement learning.}} Different from the standard RL setting, the reward function is not available in PbRL. Instead, PbRL learns the reward function $\widehat{R}$ from preferences between pairs of trajectory segments to align the human intention \cite{wilson2012bayesian}, where a trajectory segment $\sigma$ of length $L$ is defined as a sequence of states and actions $\{s_k,a_k,...,s_{k+L-1},a_{k+L-1}\}\in (\cS\times\cA)^{L}$. Given a pair of segments $(\sigma_0,\sigma_1)$, human choose which segment is preferred, i.e., $y\in\{0,1\}$ \cite{christiano2017deep}. 
\ttx{The preference label $y=1$ indicates $\sigma_1\succ\sigma_0$, while $y = 0$ indicates $\sigma_0\succ\sigma_1$, where $\sigma_i\succ\sigma_j$ denotes that segment $i$ is preferred over segment $j$.}

The preference data is stored in the dataset $\mathcal{D}_{l}$ as a tuple $(\sigma_0,\sigma_1,y)$. To obtain the reward function $\widehat{R}$ parameterized by $\psi$, the Bradley-Terry model \cite{bradley1952rank} is utilized to define a preference predictor following previous works \cite{kim2023preference,verma2024hindsight}:
\begin{equation}
    \label{eq_2}	
    P_{\psi}\left(\sigma_0\succ\sigma_1\right)=\frac{\exp{\sum_t \widehat{R}_{\psi}(s_{t}^{0},a_{t}^{0})}}{\exp{\sum_t \widehat{R}_{\psi}(s_{t}^{0},a_{t}^{0})}+\exp{\sum_t \widehat{R}_{\psi}(s_{t}^{1},a_{t}^{1})}}.
\end{equation}

Then, based on the collected preference dataset $\mathcal{D}_{l}$, the reward function $\widehat{R}_{\psi}$ can be updated through minimizing the cross-entropy loss between the predicted and the true preference labels \cite{verma2024hindsight}:
\begin{equation}
    \label{eq_3}
    \begin{aligned}
        \mathcal{L}_{R}(\psi)=-\mathbb{E}_{(\sigma_0,\sigma_1,y)\sim\mathcal{D}_{l}}\big[&y\log P_{\psi}\left(\sigma_1\succ\sigma_0\right)\\
+&(1-y)\log P_{\psi}\left(\sigma_0\succ\sigma_1\right)\big].
    \end{aligned}
\end{equation}
\ttx{Conversely, given a learned reward model, the preference label between a pair of trajectories ($\sigma_0,\sigma_1$) is determined as follows: if the probability computed by Eq. \eqref{eq_2} exceeds 0.5, trajectory 0 ($\sigma_0$) is inferred to be preferred over trajectory 1 ($\sigma_1$); otherwise, preference is assigned to $\sigma_1$.} After obtaining the reward function $\widehat{R}$, offline PbRL optimizes the policy by maximizing the expected discounted return $J(\pi, R)$ like standard offline RL.

{\textbf{Model-based reinforcement learning.}} Model-based RL learns the dynamics model to improve sample efficiency \cite{rigter2022rambo,liu2023micro,liu2023domain}. They used the offline dataset $\mathcal{D}_{\text {offline }}$ to estimate transition model $\widehat{T}(s^{\prime}|s, a)$. The transition model $\widehat{T}_{\phi}$ parameterized by $\phi$ is typically trained via maximum likelihood estimation (MLE): 
\begin{equation}
\label{eq_4}
    \mathcal{L}_{T}(\phi)=-\mathbb{E}_{(s, a,s^{\prime})\sim \mathcal{D}_{\text {offline }}}\big[\log \widehat{T}_{\phi}(s^{\prime}~|~s, a)\big].
\end{equation}
\ttx{In practical implementation, to account for the inherent uncertainty in the next-state prediction, a Gaussian distribution is employed to approximate the transition model. To further mitigate the uncertainty associated with model predictions, an ensemble of $N_T$ transition models $\{T^{i}_{\phi}=\mathcal{N}(\mu^{i}_{\phi},\sigma^{i}_{\phi})\}_{i=1}^{N_T}$ is trained. These models are then optimized using a maximum likelihood function \cite{yu2020mopo}.}
The samples are generated through $H$-step rollouts. Here, we use the trained transition to generate more trajectories to achieve data augmentation.

	\subsection{Problem Formulation}\label{problem}
The cost of collecting preference data is high since it needs real-time human feedback. Therefore, this paper aims to improve sample efficiency to learn an accurate reward model from limited preference data and guarantee agent performance. For PbRL, the ideal form is that the learned reward function $\widehat{R}$ from collected preference data can be consistent with the true reward $R^{*}$. Here, we define the function class as $\mathcal{R}=\{R:\cS \times \cA \rightarrow \mathbb{R}\}$. Then, similar to \cite{xie2021bellman,wang2021statistical}, we make the below assumption for the reward model: 
\begin{assumption}[Realizability]
    \label{ass_1}
	Let $d(s,a)\in \Delta(\cS \times \cA)$ be the arbitrary data distribution. Then, for any distribution $d(s,a)$, $\inf_{R \in \mathcal{R}}\mathbb{E}_{(s,a) \sim d(s,a)}[R^{*}(s,a)-R(s,a)]^2<\epsilon_{R}$ holds.
\end{assumption}

This assumption states that there must exist a reward function $R \in \mathcal{R}$ which can well-approximate the optimal reward function when the preference data can cover all data space. \ttx{However, preference data tend to be limited, with even fewer than a hundred in real-world scenarios.} Therefore, different from previous PbRL \cite{park2022surf,hu2024query,zhang2024flow}, this paper aims to study how to learn a better reward function from the limited dataset $\mathcal{D}_l=\{(\sigma_0^l,\sigma_1^l,y)^{(i)}\}_{i=1}^{N_l}$. To broaden the coverage of preference data, motivated by model-based offline RL methods, we train the transition model through Eq. (\ref{eq_4}) to generate more unlabeled preference data $\mathcal{D}_u=\{(\sigma_0^u,\sigma_1^u)^{(i)}\}_{i=1}^{N_u}$. \ttx{Then, the pseudo labels $\widehat{y}$ for unlabeled dataset $\mathcal{D}_u$ can be obtained through reward model trained on $\mathcal{D}_l$ \cite{park2022surf}, that is}
\begin{equation}
    \label{eq_5}	
\widehat{y}\left(\sigma_0^u,\sigma_1^u\right)= \mathbbm{1}\big[P\left(\sigma_0^u\succ\sigma_1^u; {\psi}\right)<0.5\big],
\end{equation}
\ttx{where $\mathbbm{1}(\cdot)$ is indicator function. When the probability is less than 0.5, it indicates that $\sigma_1 \succ \sigma_0$ is highly likely to hold, and the preference label is accordingly set to 1; otherwise, it is set to 0.} The reward function can be updated through the collected labeled dataset $\mathcal{D}_l$ and the generated unlabeled dataset $\mathcal{D}_u$. Then, the reward model can be optimized by minimizing
\begin{equation}
    \label{eq_6}	
    \begin{aligned}
        \mathcal{L}'_{R}(\psi)= &\mathbb{E}_{(\sigma_0^l,\sigma_1^l,y)\sim\mathcal{D}_{l}}\big[
L\left((\sigma_0^l,\sigma_1^l),y\right)\big]\\
+&\mathbb{E}_{(\sigma_0^u,\sigma_1^u)\sim\mathcal{D}_{u}}\big[
L\left((\sigma_0^u,\sigma_1^u),\widehat{y}\right)\big],
    \end{aligned}
\end{equation}
where $L((\sigma_0,\sigma_1),y)=-(1-y)\log P(\sigma_0\succ\sigma_1; {\psi})-y\log P(\sigma_1\succ\sigma_0; {\psi})$. However, the pretrained reward model may generate incorrect pseudo-labels for the unlabeled dataset, leading to noisy training and poor generalization \cite{rizve2021defense}. 
Therefore, one key question is how to design a data selection mechanism to improve prediction accuracy and guarantee training stability.

Another key aspect is the theory for offline PbRL. There are very few algorithms specifically designed for offline PbRL with strong guarantees, including a generalization bound for the reward model and a safe improvement guarantee for policy learning. Although Zhan \e \cite{zhan2024provable} established the lower bounds of policy improvement under partial coverage for offline PbRL, the general reward functions are defined over the whole trajectory rather than just state-action pairs, and the generalization bound of the reward model is not considered. Therefore, another key question is to develop a systematic theory for offline PbRL based on state-action pairs, including a generalization bound for the reward model and an improvement guarantee for policy learning.

	\section{Reward Learning} \label{reward}
The reward learning involves two stages: pretraining based on collected labeled data $\mathcal{D}_l$ and updating based on $\mathcal{D}_l$ and unlabeled data $\mathcal{D}_u$ during policy learning, which is essentially semi-supervised learning \cite{berthelot2019mixmatch}.
\ttx{However, the generated preference labels may contain errors, which can degrade the performance of the reward model. Therefore, this section focuses on designing a data selection function $f(\sigma_0^{u},\sigma_1^{u})$ to ensure the quality of the generated preference data and explains the factors influencing the generalization capability of the reward model.}

\subsection{Uncertainty-aware Pseudo-label Selection}
Motivated by the previous pseudo-labeling work \cite{rizve2021defense}, we select data from the unlabeled dataset $\mathcal{D}_u$ according to two principles: confidence and uncertainty. The data with high confidence and low uncertainty can be chosen for reward training. High confidence refers to the pre-trained reward model discriminating the preference of two trajectories into $\widehat{y}$ with high probability $p(\sigma_0^{u},\sigma_1^{u},\widehat{y})$. Low uncertainty refers that the $N_R$ reward models (model ensembles) predicts with small variance $\tau(\sigma_0^{u},\sigma_1^{u},N_R)$. The probability confidence $p(\sigma_0^{u},\sigma_1^{u},\widehat{y})$ and uncertainty variance $\tau(\sigma_0^{u},\sigma_1^{u},N_R)$ can be denoted as
\begin{equation}
    \label{eq_7}
    \begin{aligned}
        &p(\sigma_0^{u},\sigma_1^{u},\widehat{y})= (1-\widehat{y}) \bar{P}_{\psi}\left(\sigma_0^u\succ\sigma_1^u\right)+\widehat{y}\bar{P}_{\psi}\left(\sigma_1^u\succ\sigma_0^u\right)\\
        &\tau(\sigma_0^{u},\sigma_1^{u},N_R)= \text{{\texttt{Std} }}\big\{\bb{P}_{\psi}\left(\sigma_0^u\succ\sigma_1^u\right)\big\},
    \end{aligned}
\end{equation}
where $N_R$ is the number of reward model, \texttt{Std} $\{\bb{P}(\cdot)\}$ denotes the variance of output probability of $N_R$ reward models, and $\Bar{P}(\cdot)$ is the mean probability of $N_R$ reward models. Therefore, according to high confidence and low uncertainty, the $f(\sigma_0^{u},\sigma_1^{u})$ can be denoted as:
\begin{equation}
\label{eq_8}
f(\sigma_0^{u},\sigma_1^{u})=\mathbbm{1}\big[p(\sigma_0^{u},\sigma_1^{u},\widehat{y})>\kappa_{p}\big]\cdot \mathbbm{1}\big[\tau(\sigma_0^{u},\sigma_1^{u})<\kappa_{\tau}\big],
\end{equation}
where $\kappa_{p}$ and $\kappa_{\tau}$ are thresholds of confidence and uncertainty respectively. \ttx{Here, $f(\sigma_0^{u},\sigma_1^{u})=1$ denotes that the generated data pair $(\sigma_0^{u},\sigma_1^{u})$ is selected for inclusion, whereas $f(\sigma_0^{u},\sigma_1^{u})=0$ indicates its exclusion.} 

Intuitively, more rigorous data selection criteria, characterized by higher confidence and lower variance, lead to a reduction in the number of qualified samples. While such a strategy improves the quality of the generated data, it could lead to a narrower data distribution, which might adversely affect the precision of the trained reward model. Therefore, an appropriate selection of parameters $\kappa_{p}$ and $\kappa_{\tau}$ is required to enhance the performance of the reward model.
\ttx{The unlabeled dataset selected through $f(\sigma_0^{u},\sigma_1^{u})$ is denoted as $\widetilde{\mathcal{D}}_u=\{(\sigma_0^u,\sigma_1^u)^{(i)}\}_{i=1}^{\widetilde{N}_u}$, where $\widetilde{N}_u$ is the number of generated dataset after screening. }
Then, combining Eq. \eqref{eq_6} and the designed selection mechanism, the reward model can be optimized by minimizing
\begin{equation}
    \label{eq_9}
    \begin{aligned}
        \widehat{\mathcal{L}}_{R}(\psi)= \frac{1}{\widetilde{N}_u}\sum\nolimits_{j=1}^{{N}_u} f\left(\sigma_0^{u},\sigma_1^{u}\right)^{(j)}
L\left((\sigma_0^u,\sigma_1^u)^{(j)},\widehat{y}^j\right)\\
\frac{1}{N_l}\sum\nolimits_{i=1}^{N_l}L\left(\sigma_0^l,\sigma_1^l)^{(i)},y^i\right),
    \end{aligned}
\end{equation}
where $\widetilde{N}_u$ is the number of generated dataset after screening, and $\widehat{\mathcal{L}}_{R}(\psi)$ is the empirical risk. It contains two parts: labeled loss $\widehat{\mathcal{L}}_l(\psi)$ and unlabeled loss $\widehat{\mathcal{L}}_u(\psi)$. Note that the label in the unlabeled loss is a pseudo-label, which may be different from the true label. Therefore, there is a gap between the unlabeled loss with pseudo-label $\widehat{\mathcal{L}}_u(\psi)$ and that with true label $\widehat{\mathcal{L'}}_u(\psi)$. Before bounding this gap, we first give the following assumption without loss of generality.
\begin{assumption}
    \label{ass_2}
For the pretrained reward model $\widehat{R}_{\psi}$ through the limited labeled dataset $\mathcal{D}_l=\{(\sigma_0^l,\sigma_1^l,y)^{(i)}\}_{i=1}^{N_l}$, if the pseudo label $\widehat{y}$ is defined in Eq. (\ref{eq_5}), then the pseudo-labeling error for the unlabeled dataset $\mathcal{D}_u=\{(\sigma_0^u,\sigma_1^u)^{(j)}\}_{j=1}^{N_u}$ is smaller than $\eta$, i.e., $\sum_{j=1}^{N_u}\mathbbm{1}[\widehat{y}^j\neq y^j]/N_u\leq \eta$.
\end{assumption}

\ttx{Thanks to its capacity to effectively filter out samples with low confidence and high uncertainty, the screening mechanism mitigates the adverse impact of negative samples on reward model training, thereby contributing to improved reward model performance. Consequently, the bound on the pseudo-label error rate $\eta$ for the filtered generated data is significantly lower than that for the unfiltered data.} This observation is experimentally validated in Table \ref{tab_uncertainty} and Fig. \ref{fig3}.
Then, based on the above assumption, we bound the difference between the unlabeled loss with pseudo-label $\widehat{\mathcal{L}}_u(\psi)$ and that with true label $\widehat{\mathcal{L'}}_u(\psi)$ and give the following proposition.

\begin{proposition}
\label{pro_1}
    Suppose that the loss $L((\sigma_0,\sigma_1),y))$ is bounded by $\Omega$. Then, for any $R\in\mathcal{R}$, under Assumption \ref{ass_2} the following equation holds:
    \begin{equation}
\big|\widehat{\mathcal{L}}_u(\psi)-\widehat{\mathcal{L'}}_u(\psi)\big|\leq \eta\Omega.
    \end{equation}
\end{proposition}
The proof of Proposition \ref{pro_1} can be found in Appendix A. 
This gap is mainly influenced by the terms $\eta$ and $\Omega$. In the above part, we have concluded that the $\eta$ can be significantly reduced by selecting high-confidence and low-uncertainty samples. Combining Eqs. (\ref{eq_5}) and (\ref{eq_8}), we can find that if $p(\sigma_0^{u},\sigma_1^{u},\widehat{y})$ is small, the $L((\sigma_0^u,\sigma_1^u),\widehat{y})$ would become larger. By selecting higher confidence samples, the high probability $p(\sigma_0^{u},\sigma_1^{u},\widehat{y})$ can be guaranteed, thus the $\Omega$ does not become excessively large. Moreover, through the selection mechanism $f(\sigma_0^{u},\sigma_1^{u})$, a more accurate subset of pseudo-labels can be used in reward training, which can reduce prediction error and improve training stability.

\subsection{Generalization Bound for Reward Model}
This part studies the generalization performance of the reward model. 
Before developing generalization bound, we define the expected error $\mathcal{L}_{R}(\psi)$ with respect to the reward model $R(s,a;\psi)$: $\mathcal{L}_{R}(\psi)=\mathbb{E}_{(\sigma_0,\sigma_1,y)\sim\mathcal{D}}[L((\sigma_0,\sigma_1),y)]$, where $\mathcal{D}$ is data distribution. The reward model is trained through  minimizing the empirical error $\widehat{\mathcal{L}}_{R}(\psi)$ in Eq. (\ref{eq_9}). Through developing generalization error bound, 
we can analyze the factors that influence generalization ability.

Similar to the previous generalization bound works \cite{xie2024class}, Rademacher complexity, which measures the richness of a certain hypothesis space \cite{mohri2012new}, is introduced first. The definition is given below:

\begin{definition}[Empirical Rademacher complexity]\label{def_1}
	Let $\mathcal{F}$ be a family of functions mapping from $\mathcal{Z}$ to $\mathbb{R}$ and $\widehat{\mathcal{S}}=\left\{z_1, \ldots, z_{N_s}\right\}$ be a fixed sample of size $N_s$ drawn from the distribution $\mathcal{S}$ over $\mathcal{Z}$. The empirical Rademacher complexity of $\mathcal{G}$ for sample $\widehat{\mathcal{S}}$ is defined as
	\begin{equation}
		\label{8}
		\widehat{\Re}_{\widehat{\mathcal{S}}}(\mathcal{F})= \mathbb{E}_{\bb{\sigma}} \big[\sup _{f \in \mathcal{F}} \frac{1}{N_s}\sum\nolimits_{i=1}^{N_s} \sigma_i f\left(z_i\right)\big],
	\end{equation}
 where $\bb{\sigma}=(\sigma_1,...,\sigma_{N})$ are independent uniform random variables taking values in $\{-1,+1\}$.
\end{definition}
\begin{definition}[Induced Reward Function Families]\label{def_2}
Given for a space $\mathcal{R}$ of reward function $R$, the induced reward function families $\Pi(\mathcal{F})$ is defined as
\begin{equation}
	\label{9}
  \Pi(\mathcal{R}) = \big\{(\sigma_0,\sigma_1)\rightarrow -\log P(\sigma_0\succ\sigma_1)~|~R\in\mathcal{R}\big\},
\end{equation}
where $P(\cdot)$ is defined in Eq. (\ref{eq_2}) and $\Pi(\mathcal{R})$ is the union of projections of $\mathcal{R}$ onto each dimension. 
\end{definition}
In general, lower Rademacher complexity corresponds to better generalization performance. Then, based on the above definition and Proposition \ref{pro_1}, we develop the new generalization bound for the reward model trained through Eq. (\ref{eq_9}). The specific theorem is given below.
\begin{theorem}
\label{the_1}
   Let reward model be trained on the labeled dataset $\mathcal{D}_l=\{(\sigma_0^l,\sigma_1^l,y)^{(i)}\}_{i=1}^{N_l}$ and unlabeled dataset $\mathcal{D}_u=\{(\sigma_0^u,\sigma_1^u)^{(i)}\}_{i=1}^{N_u}$. Then, for any $\delta>0$, with probability at least $1-\delta$, under Assumption \ref{ass_1} and \ref{ass_2} the following holds for any reward function $R \in \mathcal{R}$,
   \begin{equation*}
    \mathcal{L}_{R}(\psi)\leq \widehat{\mathcal{L}}_{R}(\psi)+\eta\Omega+4\widehat{\mathfrak{R}}_{{\widehat{\mathcal{D}}}}\big(\Pi(\mathcal{R})\big)+3\sqrt{\frac{\log (2 / \delta)}{2(N_l+N_u)}},
\end{equation*}
where $\widehat{\mathcal{D}}$ is the input combination of labeled and unlabeled dataset, i.e. $\{(\sigma_0,\sigma_1)^{(i)}\}_{i=1}^{N_l+N_u}$.
\end{theorem}

The proof of Theorem \ref{the_1} can be found in Appendix B. 
\ttx{Theorem \ref{the_1} indicates that the expected error $\mathcal{L}_{R}(\psi)$ is bounded by the empirical error $\widehat{\mathcal{L}}_{R}(\psi)$, the pseudo-label error $\eta\Omega$, the Rademacher complexity, and a constant term. While the constant term diminishes as the amount of unlabeled data increases, inaccurate labels can lead to training instability. According to Proposition \ref{pro_1}, a selection mechanism $f(\sigma_0^{u},\sigma_1^{u})$ can be employed to reduce both the empirical error and the pseudo-label error, thereby achieving better generalization performance, that is, a tighter upper bound. Conversely, without such a screening mechanism, the pseudo-label error rate would rise, resulting in a larger upper bound for the expected error of the reward model and ultimately degrading its performance.} Moreover, Theorem \ref{the_1} has generality and can apply to methods that train a reward model using pseudo-labels. 




	\section{Policy Learning}\label{policy}
To solve the problem of high cost for labeling preference, we propose a off\textbf{L}ine pr\textbf{E}ference-b\textbf{A}sed RL with high \textbf{S}ample \textbf{E}fficiency algorithm (\texttt{LEASE}). This section aims to describe the details of \texttt{LEASE} and establish the theory for a safe policy improvement guarantee. \ttx{In this paper, we directly use the proposed offline RL algorithms, such as Conservative Q-Learning (CQL) \cite{kumar2020conservative} and Implicit Q-Learning (IQL) \cite{iql}, to perform policy learning, and do not improve the offline RL algorithm itself.}

\subsection{The Implementation Details of LEASE}

Most offline RL algorithms are based on Eq. (\ref{eq_1}). They proposed various algorithms to solve the problem of distribution shift. In an offline PbRL setting, the reward is unknown to the agent, and the accuracy of the reward model directly influences the performance of the policy. Moreover, the unlabeled data are generated through the current policy interaction with the learned transition model. The learned policy affects the distribution of the generated unlabeled preference dataset. In practical implementation, the policy, $Q$-function, reward and transition model are parameterized by related neural networks $\pi_\theta$, $Q_\omega$, $R_{\psi}$ and $T_{\phi}$, respectively.

Algorithm \ref{alg1} gives pseudo-code for \texttt{LEASE}. The goal of \texttt{LEASE} is to learn a better reward model $R_{\psi}$ from a smaller labeled preference dataset $\mathcal{D}_l$ and learn a better policy $\pi_\theta$ from an offline dataset $\mathcal{D}_{\text{offline}}$ without interaction with the environment. The key to \texttt{LEASE} is how to train the reward model in the policy learning process. The proposed algorithmic framework comprises three key components: reward model training, transition model training, and policy training. \ttx{Initially, the transition model is trained on a limited offline dataset, whereas the reward model is pretrained using a small-scale preference dataset. The reward update mechanism is activated once the volume of unlabeled data in the buffer reaches its upper capacity limit. Subsequently, the policy training phase is initiated. In this phase, the transition model is periodically leveraged to generate unlabeled synthetic data, and the reward model undergoes iterative updates to further boost its performance.}


\subsection{Safe Policy Improvement Guarantee}
Offline RL aims to guarantee $\xi$-safe policy improvement over the behavior policy $\mu$ (the policy used to collect offline dataset), \textit{i.e.} $J(\widehat{\pi},R^*)\geq J(\mu,R^*)-\xi$ \cite{kumar2020conservative}. For offline PbRL, we further investigate whether similar policy improvement guarantees can be established. \ttx{Unlike offline RL, the reward function in PbRL is unknown, and the learned reward function may contain errors, which introduces an additional performance gap.} Therefore, this part aims to develop a theoretical guarantee of policy improvement for \texttt{LEASE}, that is, giving the bound for $J(\mu,R^*)-J(\widehat{\pi},R^*)$. Then, based on the bounds of policy improvement, we further analyze the sample complexity for \texttt{LEASE} when the reward model is unknown. 

\begin{figure}[!t]
	\centering
	\renewcommand{\algorithmicrequire}{\textbf{Input:}}
	\renewcommand{\algorithmicensure}{\textbf{Output:}}
	\removelatexerror
\begin{algorithm}[H]
	\caption{\texttt{LEASE}: off\textbf{L}ine pr\textbf{E}ference-b\textbf{A}sed RL with high \textbf{S}ample \textbf{E}fficiency}
	\label{alg1}
	\begin{algorithmic}[1] 
    
\REQUIRE offline dataset $\mathcal{D}_{\text{offline}}$, limited labeled preference dataset $\mathcal{D}_l$, critic $Q_{\omega}$,  policy $\pi_\theta$, transition model ${T}_\phi$, and reward model $R_{\psi}$. 

\ENSURE  reward $\widehat{R}$ and policy $\widehat{\pi}$ learned by \texttt{LEASE}.

\STATE \textbf{Initialization:} Randomly initializing all networks and generated unlabeled dataset $\mathcal{D}_u=\varnothing$.

\STATE Train an ensemble transition model $\{{T}_\phi^i\}_{i=1}^{N_T}$ on $\mathcal{D}_{\text{offline} }$ according to Eq. (\ref{eq_4}).
\STATE Pre-train an ensemble reward model $\{{R}_\psi^i\}_{i=1}^{N_R}$ on $\mathcal{D}_l$ according to Eq. (\ref{eq_3}).
  \FOR {$t=1,2,\cdots, N_{\text{iter}} $}
  \IF {$t$ \% \textit{rollout frequency} $=0$}
        \STATE Generate data by transition model ${T}_\phi$, and add transition trajectory into $\mathcal{D}_u$.
\ENDIF
\IF {$t$ \%  \textit{reward update frequency} $=0$}
\STATE Generate pseudo label for unlabeled dataset $\mathcal{D}_u$, and screen unlabeled data by Eq. (\ref{eq_8}).
\STATE Train the reward model based on $\mathcal{D}_l$ and $\mathcal{D}_u$ according to Eq. (\ref{eq_9}).
\ENDIF
\STATE Train policy using offline RL algorithms.
 \ENDFOR
\end{algorithmic}
\end{algorithm}
\end{figure}

Before deriving the bounds of policy improvement, we first define a new single-policy concentrability coefficient for PbRL based on the state-action pair. This coefficient can quantify the discrepancy between the offline distribution and the visitation distribution of policy $\pi$. 

\begin{definition}[Concentrability coefficient for PbRL]\label{def_3}
Concentrability coefficient $\mathscr{C}_{\mathcal{R}}(\pi)$ is used to measure how well reward model errors transfer between offline data distribution $d_{T}^{\mu}$ and the visitation distribution $d_{T}^{\pi}$ under transition $T$ and policy $\pi$, defined as
\begin{equation}
\label{eq_14}
    \mathscr{C}_{\mathcal{R}}(\pi)=\Bigg|\frac{\mathbb{E}_{(s,a) \sim d_{T}^{\pi}}\big[R^*(s,a)-\widehat{R}(s,a)\big]}{\mathbb{E}_{(s,a) \sim d_{T}^{\mu}}\big[R^*(s,a)-\widehat{R}(s,a)\big]}\Bigg|,
\end{equation}
where $R^*(s,a)$ is the true reward model and $\hat{R}(s,a)$ is the learned reward model.
\end{definition}

This concentrability coefficient $\mathscr{C}_{\mathcal{R}}(\widehat{\pi})$ is influenced by the accuracy of the learned reward model $\hat{R}(s,a)$ and the distribution $d_{T}^{\pi}$. It can become smaller when the learned policy $\widehat{\pi}$ is close to the behavior policy $\mu$. Next, similar to \cite{geer2000empirical,zhan2024provable}, to measure the complexity of reward function class $\mathcal{R}$, we use $\epsilon$-bracketing number, which can be defined as

\begin{definition}[$\epsilon$-bracketing number]
\label{def_4}
The $\epsilon$-bracketing number $\mathcal{N}_{\mathcal{R}}(\epsilon)$ is the minimum number of $\epsilon$-brackets $(l,u)$ required to cover a function class $\mathcal{R}$, where each bracket $(l,u)$ satisfies $l(\sigma_0,\sigma_1)\leq P_{R}(\sigma_0, \sigma_1)\leq u(\sigma_0,\sigma_1)$ and $\|l-u\|_1\leq \epsilon$ for all $R\in \mathcal{R}$ and all trajectory-pairs $(\sigma_0,\sigma_1)$, and $P_{R}(\sigma_0, \sigma_1)$ is the probability that segment $\sigma_0$ is preferable to segment $\sigma_1$ defined in Eq. (\ref{eq_2}).
\end{definition}

Bracketing number $\mathcal{N}_{\mathcal{R}}(\epsilon)$ measures the complexity of function class $\mathcal{R}$ and takes $\widetilde{\mathcal{O}}(d)$ in linear reward model \cite{zhan2024provable}. Then, before giving the theorem of policy improvement, we first assume the reward class $\mathcal{R}$ is bounded, that is
\begin{assumption}[Boundedness]
    \label{ass_3}
	For any $R\in \mathcal{R}$ and any state-action pairs $(s,a)$, the equation $|R(s,a)|\leq R_{max}$ holds.
\end{assumption}

\begin{theorem}
\label{the_2}
Under Assumptions \ref{ass_1} and \ref{ass_3}, for any $\delta \in (0,1]$, the policy $\widehat{\pi}$ learned by \texttt{LEASE}, with high probability $1-\delta$, satisfies that $J(\mu,R^*)-J(\widehat{\pi},R^*)$ is upper bound by
   \begin{equation}
   \begin{aligned}
         \xi+\frac{1+\mathscr{C}_{\mathcal{R}}(\widehat{\pi})}{1-\gamma}\Bigg(\sqrt{\frac{4C}{NL^2}\log\left(\frac{\mathcal{N}_{\mathcal{R}}(1/N)}{\delta}\right)}\\
         +\tx{\sqrt{\frac{4R_{max}^2\log \big(1/\delta\big)}{NL}}}\Bigg), 
   \end{aligned}
    \end{equation}
where $C>0$ is the absolute constant defined in Lemma 2, $N$ is the size of preference dataset, and $L$ is the length of trajectory. The term $\xi$ is the performance gap depending on the offline algorithm itself.
\end{theorem}
The proof of Theorem \ref{the_2} can be found in Appendix C. 
\ttx{The performance bound for the PbRL policy learned by \texttt{LEASE} consists of two distinct components: the first term, denoted as $\xi$, stems from the inherent gap of the offline RL process itself; the second term arises from the discrepancy induced by the reward model. When the offline algorithm is fixed, the first term $\xi$ is constant. The second term depends on the performance of the learned reward model. Consequently, the proposed framework is readily compatible with other offline RL algorithms, and a tighter overall bound can be achieved by reducing the performance gap attributable to the reward model (see Proposition 3).} It can be reduced to a small value $\varpi$ with a sample complexity of
\begin{equation}
\label{eq_16}
\begin{aligned}
N=\widetilde{\mathcal{O}}\Bigg(\frac{4\big(1+\mathscr{C}_{\mathcal{R}}(\widehat{\pi})\big)^2}{\varpi^2(1-\gamma)^2}
\Bigg(&\sqrt{\frac{C\log (\mathcal{N}_{\mathcal{R}}(1/N)/\delta)}{L^2}}\\
&+\sqrt{\frac{R_{max}^2\log \big(1/ \delta\big)}{L}}\Bigg)^2\Bigg). 
\end{aligned}
\end{equation}
This sample complexity is influenced by the concentrability coefficient $\mathscr{C}_{\mathcal{R}}(\widehat{\pi})$, the complexity of the function class $\mathcal{R}$ $\log(\mathcal{N}_{\mathcal{R}}(1/N))$ and the bounded of reward model $R_{max}$ and the length of preference trajectory $L$. The performance of the reward model learned by \texttt{LEASE} can be improved compared with that of not using the selection mechanism (Theorem \ref{the_1}). Then, the gap between the true reward model and the learned reward model can be reduced, thus the concentrability coefficient $\mathscr{C}_{\mathcal{R}}(\widehat{\pi})$ can be further reduced. Moreover, \texttt{LEASE} can learn an accurate reward model through data augmentation under fewer $N_l$ preference data ($N_l<N$). Thus, the bounds of the policy performance gap learned by \texttt{LEASE} can become tighter with fewer data.
	\section{Experiments}\label{experiments}
This section gives the details of experiments and presents comprehensive experimental evaluations of the proposed algorithm, including the accuracy assessment of reward model learning, performance analysis of policy learning, and detailed parameter sensitivity studies. We focus on answering the following questions: 
\begin{itemize}
\item $\mathbf{Q_1}$: How well does the accuracy of the reward model through data augmentation under fewer preference data?
\item $\mathbf{Q_2}$: How well does \texttt{LEASE} perform compared with the offline PbRL baseline in standard benchmark tasks? 
\item $\mathbf{Q_3}$: How do the proposed selection mechanism and parameter settings affect the performance of reward and policy? 
\end{itemize}

We answer these questions on two distinct domains (mujoco and adroit), covering several locomotion and manipulation tasks, within the D4RL benchmark \cite{fu2020d4rl}. The code for \texttt{LEASE} is available at \href{https://github.com/xiaoyinliu0714/LEASE}{Code}.

\subsection{Experiment Setup}
\begin{figure}
	\centering
	\includegraphics[width=0.48\textwidth]{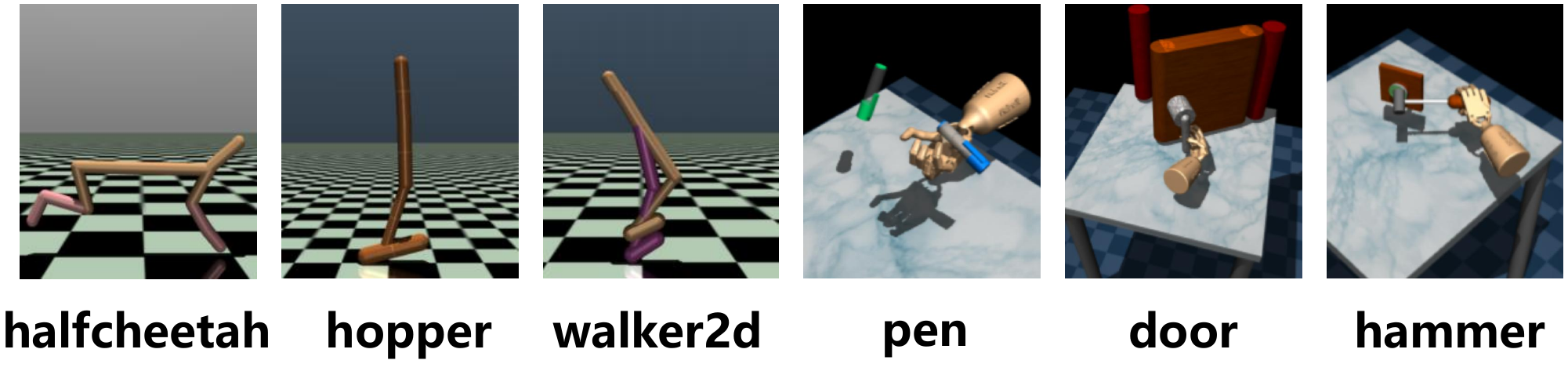}
	\caption{The description of Mujoco tasks (locomotion tasks) and Adroit tasks (manipulation tasks).}. 
	\label{figA1}
\end{figure}

\subsubsection{Experimental Datasets} The offline dataset and preference dataset of experiments originate from \cite{fu2020d4rl} and \cite{yuanuni}, respectively. The offline datasets are based on the D4RL dataset \cite{fu2020d4rl}. \ttx{We select two types of datasets for Mujoco tasks: medium and medium-expert, and human and expert for Adroit tasks.} The difference between datasets in a certain task lies in the policies they collect. The Mujoco tasks include halfcheetah-v2, hopper-v2, and walker2d-v2. The Adroit tasks include pen-v1, door-v1, and hammer-v1. 
Fig. \ref{figA1} depicts the six tasks for the D4RL benchmark. \ttx{The objective of the three Mujoco tasks is to control the robot's joints to achieve faster and more stable motion, with action dimensions ranging from 3 to 8, and the reward function is defined in the form of continuous values. In contrast, the Adroit task involves controlling a simulated Shadow Hand robot with an action dimension of 24, which is a high-dimensional robotic manipulation task characterized by sparse rewards—only a few specific reward values are provided.}

\ttx{The preference dataset is from the previous work \cite{yuanuni}, where they provided the baseline for offline PbRL based on $2000$ preference data labeled by two types (We denote this method as \texttt{URLHF}). One is crowd-sourced labels obtained through crowd-sourcing, and the other is synthetic labels based on ground-truth rewards.} To test \texttt{LEASE} performance for two types of preference data, the labels of preference are from human-realistic feedback for MuJoco tasks, and ground truth reward for Adroit tasks. This paper aims to test the performance under fewer preference data without online interaction. We take the first $100$ data of \cite{yuanuni} as $2000$ preference dataset, and denote the method without data augmentation as \texttt{FEWER}.

\subsubsection{Experimental Details} The Offline PbRL setting involves two steps: reward learning and policy learning. Here, we directly use offline RL algorithms CQL \cite{kumar2020conservative} and IQL \cite{iql} to perform policy learning. The performance of these algorithms is evaluated based on the cumulative return. For comparison, the scores are normalized between 0 (random policy score) and 100 (expert policy score) \cite{fu2020d4rl}. The normalized score $\tilde{J}$ is computed by:
\begin{equation}
	\tilde{J}=\frac{J-J_r}{J_e-J_r} \times 100,
\end{equation}
where $J_r$, $J_e$, and $J$ are the expected returns of a random policy, an expert policy, and the trained policy by offline RL algorithms, respectively. Table \ref{Base} gives the hyperparameter configuration for reward and transition training, and policy learning. The details are given below:

\begin{figure*}
	\centering
	\includegraphics[width=0.9\textwidth]{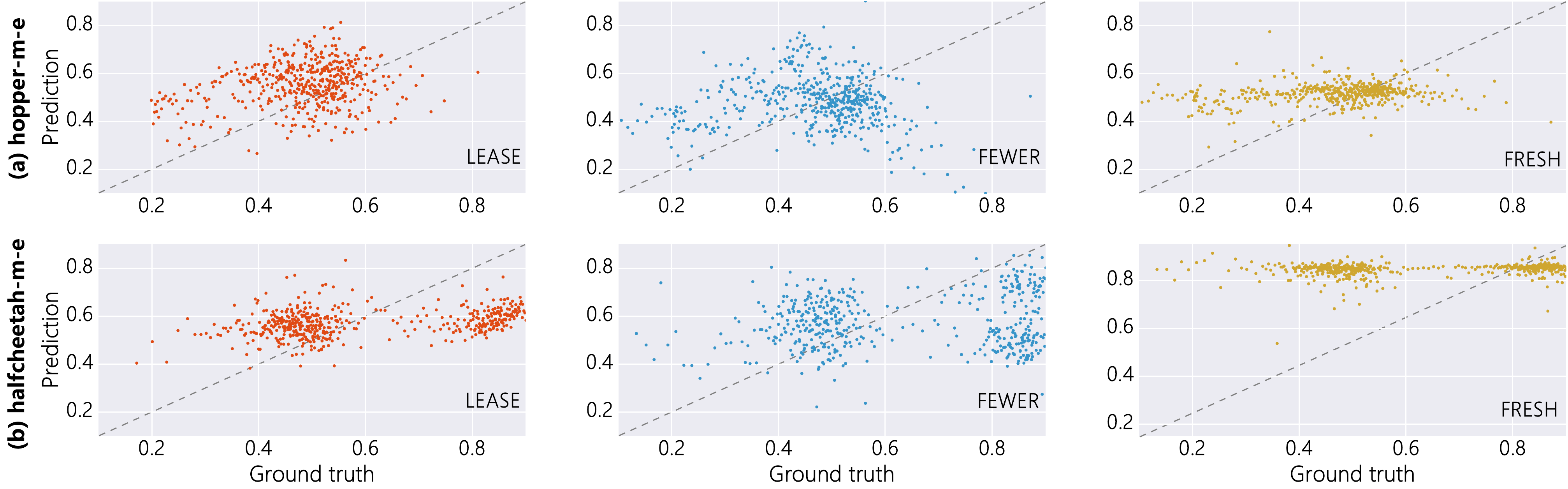}
	\caption{The comparison between prediction value by the learned rewards and their ground truths for different methods under (a) hopper-medium-expert and (b) halfcheetah-medium-expert datasets, where the value predicted by the trained reward model and ground-truth reward value are both normalized to $[0,1]$. From left to right are methods \texttt{LEASE}, \texttt{FEWER} and \texttt{FRESH}, respectively.}. 
	\label{fig3}
\end{figure*}

\begin{table}
	\centering
	\caption{Base parameter configuration of method \texttt{LEASE}.}
	\label{Base}
	\renewcommand{\arraystretch}{1.1}
		\begin{tabular}{p{0.02cm}p{0.02cm}p{0.25cm}ll}
			\toprule[1pt] 
			\multicolumn{3}{l}{\textbf{Model}}  & \textbf{Parameter}&\textbf{Value} \\
			\hline		
			\multirow{6}{*}{\rotatebox{90}{\textbf{Transition}}}&\multirow{6}{*}{\rotatebox{90}{\textbf{Model}}}&&Model learning rate & $1\times 10^{-3}$ \\
			&&&Number of hidden layers & 4 \\
			&&&Number of hidden units per layer& 200\\
			&&&Batch size & 256 \\
			&&&Number of model networks $N_T$ & 7 \\
			&&&Number of elites & 5 \\
			\hline
			
			\multirow{6}{*}{\rotatebox{90}{\textbf{Reward}}}&\multirow{6}{*}{\rotatebox{90}{\textbf{Model}}}&&Model learning rate & $3\times 10^{-4}$ \\
			&&&Number of hidden layers & 3 \\
			&&&Number of hidden units per layer& 256\\
			&&&Batch size (Pretrain) & 256(64) \\
			&&&Number of model networks $N_R$ & 3 \\
                &&&Number of labeled dataset $N_l$ & 100 \\
                \hline
                
			\multirow{6}{*}{\rotatebox{90}{\textbf{Policy}}}&\multirow{6}{*}{\rotatebox{90}{\textbf{Model}}}
                &&Actor learning rate (IQL) & $1(3)\times 10^{-4}$ \\
                &&&Critic learning rate & $3\times 10^{-4}$ \\
			&&&Number of hidden layers(IQL) & 3(2) \\
			&&&Number of hidden units per layer& 256\\
			&&&Batch size & 256 \\
			\bottomrule[1pt]
		\end{tabular}
\end{table}



\textbf{Reward and transition model training:}  Similar to \cite{yu2020mopo}, we train an ensemble of $7$ transition models and select the best $5$ models. Each model consists of a $4$-layer feed-forward neural network with $200$ hidden units. The model training employs maximum likelihood estimation with a learning rate of $1\times 10^{-3}$ and Adam optimizer. \ttx{Following \cite{yuanuni}, we train an ensemble of $3$ reward models. Each reward model includes a $3$-layer feed-forward neural network with $256$ hidden units. The training of the reward model uses cross-entropy loss with a learning rate of $3\times 10^{-4}$ and the Adam optimizer.}

\textbf{Policy optimization:} The offline algorithms include CQL \cite{kumar2020conservative} and IQL \cite{iql}. In CQL algorithm, the actor network $\pi_{\theta}$ and critic network $Q_{\omega}$ adopt a $3$-layer feed-forward neural network with learning rate of $1\times 10^{-4}$ and $3\times 10^{-4}$, respectively. The \texttt{cql weight} and \texttt{temperature} are set as $5.0$ and $1.0$ for all tasks, respectively. In the IQL algorithm, the actor network and critic network adopt a $2$-layer neural network with learning rate of $3\times 10^{-4}$ and $3\times 10^{-4}$, respectively. Similar to \cite{iql}, the \texttt{expectile} and \texttt{temperature} are set as $0.7$ and $3.0$ for mujoco tasks and $0.8$ and $3.0$ for adroit tasks. 

\textbf{Parameters choice:} In the reward modeling aspect, the selection mechanism $f(\sigma_0^{u},\sigma_1^{u})$ is designed to enhance the quality of generated data, thereby improving the reward model's performance under limited preference data. Two critical parameters require careful selection for \texttt{LEASE}: the confidence threshold $\kappa_p(\sigma_0^{u},\sigma_1^{u},\widehat{y})$ and the uncertainty threshold $\kappa_\tau(\sigma_0^{u},\sigma_1^{u},N_R)$. These parameters are chosen from two sets: $\kappa_p\in \{0.85,0.95\}$ and $\kappa_\tau\in \{0.05, 0.08\}$. For confidence threshold choice, we set $\kappa_p$ as $0.85$ and $0.95$ for locomotion tasks (mujoco domain) and manipulation tasks (adroit domain), respectively. For the uncertainty threshold choice, we set $\kappa_\tau$ as $0.05$ and $0.08$ for CQL and IQL algorithms, respectively. For detailed techniques on parameter selection, please refer to Section \ref{Parameter}. 
\ttx{Note that RL typically entails extensive parameter fine-tuning tailored to specific tasks. In this paper, the primary focus is placed on proposing an efficient preference-based RL framework, rather than on pursuing performance gains via parameter tuning. As such, the parameters employed in this study do not necessarily represent the optimal configuration.}


\begin{figure}
	\centering
	\includegraphics[width=0.43\textwidth]{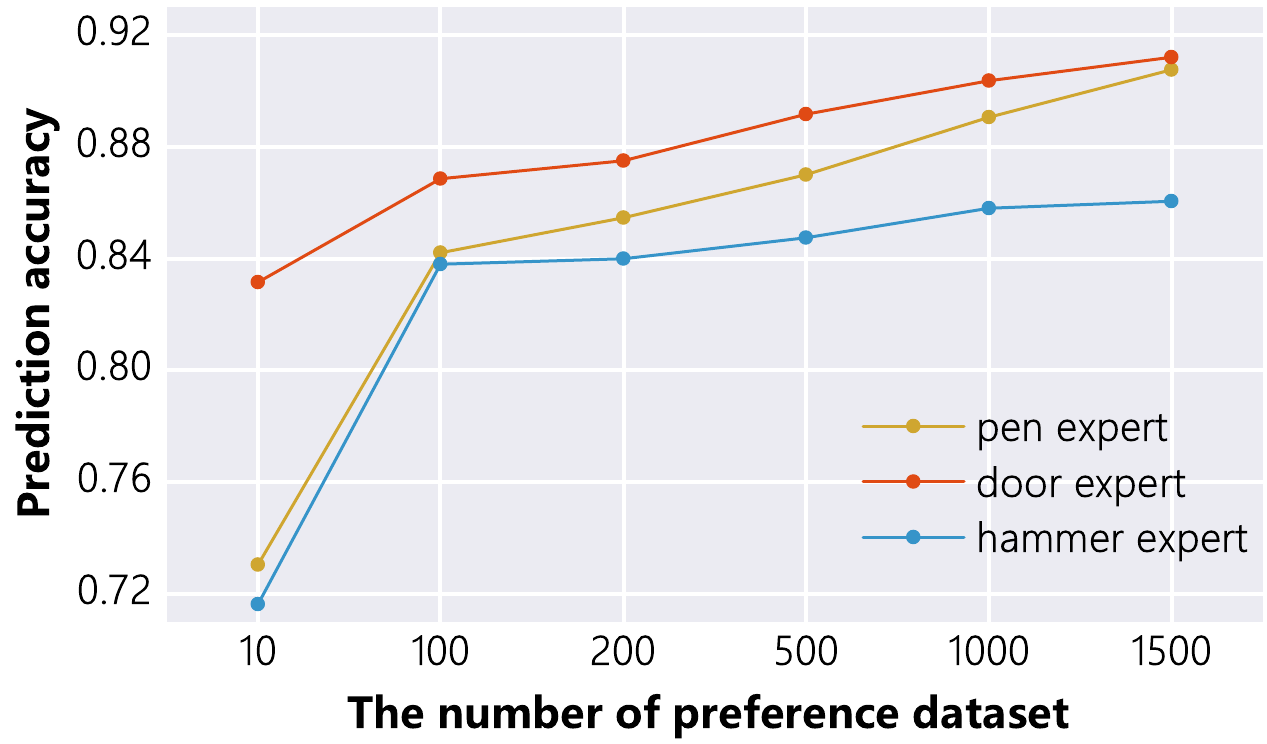}
	\caption{The relationship between pseudo-labeling accuracy and the number of preference dataset $N_l$.}. 
	\label{fig2}
\end{figure}

\begin{table*}
\caption{Results for the D4RL tasks during the last 5 iterations of training averaged over 3 seeds. $\pm$ captures the standard deviation over seeds. Bold indicates the performance within 2\% of the best-performing algorithm for each offline algorithm.}
 \label{d4rl}
	\renewcommand{\arraystretch}{0.8}
	\centering
	\resizebox{0.9\textwidth}{!}
	{
		\begin{tabular}{c|ccc|ccc}
			\toprule[1pt] 
			\multirow{2}{*}{\textbf{Task Name}} & \multicolumn{3}{c|}{\textbf{CQL} \cite{kumar2020conservative}}&\multicolumn{3}{c}{\textbf{IQL} \cite{iql}}\\
   & \texttt{\textbf{URLHF}}&\texttt{\textbf{FEWER}} & \texttt{\textbf{LEASE}} & \texttt{\textbf{URLHF}}&\texttt{\textbf{FEWER}} & \texttt{\textbf{LEASE}}  \\
\midrule
walker2d-m&  $76.0\pm0.9$ & $77.4\pm0.6$ & $\mathbf{78.4\pm0.9}$ 
            & $\mathbf{78.4\pm0.5}$ & $71.8\pm0.8$ & $73.0\pm2.1$\\
            
walker2d-m-e&  $92.8\pm22.4$ & $77.7\pm0.3$ & $\mathbf{98.6\pm 18.1}$ 
& $\mathbf{109.4\pm0.1}$ & $105.2\pm3.1$ & $\mathbf{108.1\pm0.5}$\\

hopper-m&  $54.7\pm3.4$ & $55.8\pm2.8$ & $\mathbf{56.5\pm0.6}$ 
& $50.8\pm4.2$ & $\mathbf{56.7\pm2.6}$ & $\mathbf{56.0\pm 0.5}$\\

hopper-m-e& $\mathbf{57.4\pm4.9}$ & $53.6\pm0.9$ & $56.4\pm0.8$ 
& $\mathbf{94.3\pm7.4}$ & $56.0\pm1.4$ & $55.9\pm1.9$\\

halfcheetah-m&  $\mathbf{43.4\pm 0.1}$ & $\mathbf{43.5\pm0.1}$ & $\mathbf{43.5\pm0.1}$ 
& $\mathbf{43.3\pm0.2}$ & $42.2\pm0.1$ & $\mathbf{43.0\pm0.3}$\\

halfcheetah-m-e&  $\mathbf{62.7\pm7.1}$ & $48.3\pm0.7$ & $53.2\pm3.1$ 
& $\mathbf{91.0\pm2.3}$ & $59.1\pm4.9$ & $60.2\pm1.2$\\
\midrule
\textbf{Mujoco Average} &  $\mathbf{64.5\pm6.5}$ & $59.4\pm0.9$ & $\mathbf{64.4\pm4.0}$ 
& $\mathbf{77.9\pm2.5}$ & $65.2\pm2.2$ & $66.4\pm1.1$ \\
   
\midrule
pen-human&  $\mathbf{9.8\pm14.1}$ & $0.5\pm3.0$ & $3.8\pm4.6$ 
& $50.2\pm15.8$ & $67.3\pm10.0$ & $\mathbf{75.6\pm3.3}$\\

pen-expert&  $\mathbf{138.3\pm5.2}$ & $128.1\pm0.7$ & $130.6\pm1.8$ 
& $\mathbf{132.9\pm4.6}$ & $104.1\pm12.9$ & $113.8\pm6.3$\\

door-human&  $\mathbf{4.7\pm5.9}$ & $0.2\pm1.0$ & $\mathbf{4.7\pm8.8}$ 
& $3.5\pm3.2$ & $4.0\pm2.5$ & $\mathbf{5.7\pm0.8}$\\

door-expert&$\mathbf{103.9\pm0.8}$ & $103.0\pm0.9$& $\mathbf{103.2\pm 0.7}$ 
& $\mathbf{105.4\pm0.4}$ & $104.5\pm0.6$ & $\mathbf{105.2\pm0.2}$\\

hammer-human&  $\mathbf{0.9\pm0.3}$ & $0.3\pm0.0$ & $0.3\pm0.0$ 
& $\mathbf{1.4\pm1.0}$ & $1.2\pm0.2$ & $\mathbf{1.4\pm0.9}$\\

hammer-expert&  $120.2\pm6.8$ & $124.1\pm2.1$ & $\mathbf{126.3\pm 1.2}$ 
& $\mathbf{127.4\pm0.2}$ & $125.2\pm2.3$ & $\mathbf{126.3\pm0.1}$\\
\midrule
\textbf{Adroit Average} &$\mathbf{63.0\pm5.5}$ & $59.4\pm1.3$ & $61.5\pm2.9$ 
& $\mathbf{70.1\pm4.2}$ & $67.1\pm4.8$ & $\mathbf{71.3\pm1.9}$ \\
\bottomrule[1pt]
		\end{tabular}
	}
\end{table*}

\subsection{Results for Reward Model}\label{1}
\subsubsection{The effect of data augmentation on the performance of the reward model} To answer question $\mathbf{Q_1}$, we compare the accuracy of the reward model before and after data augmentation under a smaller preference dataset. Fig. \ref{fig3} shows the comparison results between prediction and ground truth of reward, where the predicted and true rewards are both normalized to $[0,1]$. We randomly sample $500$ data points from unlabeled datasets that are not seen in the training stage for evaluation. This figure indicates that the linear relationship between reward predicted by \texttt{LEASE} and ground truth is better than that of the other two methods, where the method not using a selecting mechanism is denoted as \texttt{FRESH}. \ttx{The reward model learned by the method \texttt{FRESH} exhibits a notably narrow prediction range, and its accuracy significantly drops compared with that of \texttt{FEWER}. This is because the data generated by \texttt{FRESH} are not filtered by $f(\sigma_0,\sigma_1)$, leading to substantial errors in data labeling and consequently causing training collapse.
These results validate that the performance of a reward model can be enhanced through data augmentation and selection mechanisms, with the design of the selection mechanism having a more pronounced impact on the model’s overall performance.}

\begin{table}
\caption{ Comparison results of pseudo-labeling accuracy between using the selection mechanism (Ours), only using confidence (No-U), only using uncertainty (No-C), and not using the selection mechanism (No-C-U).}
 \label{tab_uncertainty}
	\renewcommand{\arraystretch}{1.1}
	\centering
	\resizebox{1\linewidth}{!}
	{
		\begin{tabular}{c|cccc}
			\toprule[1pt] 
           \textbf{Task Name}& \textbf{Ours}&\textbf{No-U} & \textbf{No-C}& \textbf{No-C-U}\\
\midrule
pen-expert & $\mathbf{87.3\pm0.5}$ & $85.9\pm0.8$ & $86.7\pm0.6$ & $86.8\pm0.5$\\
door-expert & $\mathbf{89.3\pm0.6}$ & $87.8\pm0.2$ & $86.0\pm0.0$ & $85.9\pm0.8$\\
hammer-expert & $\mathbf{85.5\pm0.2}$ & $84.4\pm0.4$ & $82.5\pm1.2$ & $83.6\pm0.5$\\
\midrule
\textbf{Average} & $\mathbf{87.4\pm0.4}$&${86.0\pm0.5}$ &${85.1\pm0.6}$&${85.4\pm0.6}$\\
\bottomrule[1pt]
\end{tabular}
	}
\end{table}

\ttx{\subsubsection{The effect of the selection mechanism for pseudo-label accuracy} 
This paper introduces a novel screening mechanism designed to select unlabeled preference data samples characterized by both high confidence and low uncertainty, thereby improving pseudo-label accuracy. The pseudo-labels are generated by a preference model derived from a pre-trained reward model, as defined in Eq. \eqref{eq_2}. The performance of this approach is evaluated in terms of pseudo-label accuracy across all preference datasets. As shown in Table \ref{tab_uncertainty}, the proposed screening mechanism—which simultaneously applies confidence and uncertainty thresholds—is compared against three baseline conditions: employing only the confidence threshold (No-U), employing only the uncertainty threshold (No-C), and using no screening mechanism (No-C-U).}

\ttx{Table \ref{tab_uncertainty} demonstrates that the proposed screening mechanism substantially enhances the pseudo-label accuracy, \textit{i.e.}, it reduces the error $\eta$ defined in Assumption \ref{ass_2}. For the Pen, Door, and Hammer experts, the pseudo-label accuracy is improved by 4.5\%, 2.8\%, and 6.9\%, respectively. Furthermore, we analyze the influence of the amount of preference data on the preference model. Fig. \ref{fig2} illustrates the relationship between the prediction accuracy of the preference model—without data augmentation—and the quantity of preference data $N_l$. Here, we use the final 200 preference samples (out of a total of 2000) that were not observed during training as the evaluation dataset. The results indicate that as the amount of preference data $N_l$ increases, the prediction accuracy of the preference model exhibits an upward trend. This observation also corroborates Theorem \ref{the_1}, which states that increasing the number of samples $N_l$ reduces the generalization error.}


\subsection{Results for Policy Performance}

\begin{table*}
\caption{The comparison results between the performance using the selection mechanism and that not using it. The latter method is denoted as \texttt{FRESH}. $\uparrow$ denotes the improvement of performance.}
 \label{select}
	\renewcommand{\arraystretch}{0.8}
	\centering
	\resizebox{0.9\textwidth}{!}
	{
		\begin{tabular}{p{0.45cm}c|ccc|ccc}
			\toprule[1pt] 
	
   \multicolumn{2}{c|}{\textbf{Task Name}}&walker2d-m& hopper-m & halfcheetah-m & walker2d-m-e &hopper-m-e & halfcheetah-m-e \\
\midrule
\multirow{2}{*}{\textbf{CQL}}&\texttt{\textbf{{FRESH}}} & $76.0\pm1.3$ & $54.4\pm 1.6$ & $43.4\pm 0.1$ 
& $77.4\pm 2.2$ & $55.0\pm 0.4$ & $49.9\pm 0.9$\\

&\texttt{\textbf{{LEASE}}} & $\mathbf{78.4 \uparrow (3.3\%)}$ & $\mathbf{56.5 (\uparrow4.0\%)}$ & $\mathbf{43.5\uparrow (0.6\%)}$ 
&$\mathbf{98.6\uparrow (27.4\%)}$ & $\mathbf{56.4\uparrow (2.6\%)}$ & $\mathbf{53.2\uparrow (6.6\%)}$\\

\midrule
\multirow{2}{*}{\textbf{IQL}}&\texttt{\textbf{{FRESH}}} &  $72.7\pm 4.1$ & $53.5\pm 0.8$ & $42.5\pm 0.1$
& $105.3\pm 1.0$ & $53.9\pm 0.1$ & $54.5\pm 2.2$\\

&\texttt{\textbf{{LEASE}}}& $\mathbf{73.0\uparrow (0.4\%)}$ & $\mathbf{56.0\uparrow (4.5\%)}$ & $\mathbf{43.0\uparrow (1.3\%)}$
& $\mathbf{108.1\uparrow (2.7\%)}$ & $\mathbf{55.9\uparrow (3.7\%)}$ & $\mathbf{60.2\uparrow (10.5\%)}$\\

\bottomrule[1pt]
\end{tabular}
}
\end{table*}

\subsubsection{The comparison results on benchmark tasks} To answer question $\mathbf{Q_2}$, here, we evaluate \texttt{LEASE} performance based on CQL \cite{kumar2020conservative} and IQL \cite{iql} two offline algorithms on six types of tasks and 12 datasets in total. 
Table \ref{d4rl} presents a comparative analysis of he results of the average normalized score with standard deviation under three distinct scenarios: 1) performance with extensive preference data (\texttt{URLHF}), 2) performance with limited data without employing data augmentation techniques (\texttt{FEWER}), and 3) performance with limited data while incorporating data augmentation (our proposed method \texttt{LEASE}). In this table, m and m-e represent medium and medium-expert, respectively. The results of the offline PbRL algorithm under sufficient feedback (\texttt{URLHF}) are obtained from the paper \cite{yuanuni}. As demonstrated in Table \ref{d4rl}, \texttt{LEASE} exhibits significantly superior performance compared to \texttt{FEWER}. This empirical evidence substantiates that data augmentation can effectively enhance the agent's capability. 

Furthermore, compared to the performance achieved with extensive preference data \texttt{URLHF}, \texttt{LEASE} attains comparable results in most cases. \ttx{However, the proposed method exhibits slightly inferior performance on certain specific tasks, such as the hopper-medium-expert and halfcheetah-medium-expert datasets. This limitation primarily stems from the overly narrow distribution of the expert datasets, which consequently constrains the distribution of the generated preference data. As a result, data augmentation fails to effectively enhance the performance of the reward model, thereby limiting the potential improvement of the agent's performance.} However, overall, the results in Table \ref{d4rl} can validate the content of Theorem \ref{the_2}, which shows that \texttt{LEASE} can reduce the performance gap caused by the reward model and improve the agent's performance with fewer preference data. 

\begin{table*}
\caption{The comparison results for the D4RL tasks under different number of preference data $N_l$. $\pm$ captures the standard deviation over seeds. Bold indicates the highest score.}
 \label{num}
	\renewcommand{\arraystretch}{0.8}
	\centering
	\resizebox{0.85\linewidth}{!}
	{
		\begin{tabular}{ccc|ccc}
			\toprule[1pt] 
			 \multicolumn{3}{c|}{\textbf{CQL} \cite{kumar2020conservative}}&\multicolumn{3}{c}{\textbf{IQL} \cite{iql}}\\
   {\textbf{Task Name}} & $N_l=20$ & $N_l=100$ & {\textbf{Task Name}} & $N_l=20$ & $N_l=100$  \\
\midrule
walker2d-medium&  $77.7\pm1.3$ & $\mathbf{78.4\pm0.9}$ &
pen-human&  $71.9\pm7.4$ & $\mathbf{75.6\pm3.3}$    \\
            
walker2d-medium-expert&  $98.0\pm18.6$ & $\mathbf{98.6\pm 18.1}$ &
pen-expert & $102.5\pm12.8$ & $\mathbf{113.8\pm6.3}$\\

hopper-medium&  $\mathbf{56.8\pm1.5}$ & ${56.5\pm0.6}$ &
door-human& $4.4\pm1.2$ & $\mathbf{5.7\pm0.8}$\\

hopper-medium-expert & $54.5\pm1.2$ & $\mathbf{56.4\pm0.8}$ &
door-expert & $\mathbf{105.2\pm0.1}$ & $\mathbf{105.2\pm0.2}$\\

halfcheetah-medium&  $43.4\pm0.4$ & $\mathbf{43.5\pm0.1}$ &
hammer-human& $1.2\pm0.4$ & $\mathbf{1.4\pm0.9}$\\

halfcheetah-medium-expert & $51.0\pm0.8$ & $\mathbf{53.2\pm3.1}$ &
hammer-expert & $\mathbf{126.4\pm0.1}$ & $126.3\pm0.1$\\

\midrule
{\textbf{Mujoco Average}} &  $63.6\pm4.0$ & $\mathbf{64.4\pm4.0}$ 
& {\textbf{Adroit Average}} & $68.6\pm3.7$ & $\mathbf{71.3\pm1.9}$  \\
\bottomrule[1pt]
\end{tabular}
}
\end{table*}

\subsubsection{The effect of selection mechanism on agent performance} To answer question $\mathbf{Q_3}$, we compare the agent performance between the proposed method \texttt{LEASE} and the method not using the selection mechanism (We denote this method as \texttt{FRESH}). Fig. \ref{fig3} shows that the performance of the reward model can be significantly improved by using the selection mechanism. Table \ref{select} compares the results between \texttt{LEASE} and \texttt{FRESH} under three locomotion tasks based on CQL and IQL offline RL algorithms. This table indicates that the application of the selection mechanism $f(\sigma_0,\sigma_1)$ can efficiently improve agent performance, and the average performance of the policy learned by \texttt{LEASE} can be improved by $5.6\%$ compared with not using the selection mechanism. These results also further validate the effectiveness and advantages of the designed selection mechanism.

\subsubsection{The effect of the number of preference data \texorpdfstring{$N_l$}{Nl} for agent performance} Fig. \ref{fig2} indicates that the number of preference data $N_l$ influences the accuracy of the reward model directly. \texttt{URLHF} \cite{yuanuni} trains reward model with $2000$ preference dataset. This paper aims to achieve competitive performance with limited preference data, that is, achieve comparable performance with \texttt{URLHF} under a smaller preference dataset. We further investigate the impact of the number of preference data on the agent's performance. Table \ref{num} compares the agent's performance when the number of preference datasets is $20$ and $100$, using the CQL algorithm for MuJoCo tasks and the IQL algorithm for Adroit tasks. This table indicates \texttt{LEASE} can still improve agent performance under a very small preference dataset. \ttx{It is noteworthy that the performance gap between using 100 preference datasets and 20 preference datasets is not substantial. This is primarily because the rewards in the D4RL dataset are relatively simple; a small amount of preference data is sufficient to learn a reasonably accurate reward model. Coupled with data augmentation, the performance of the reward function can be further improved, thereby enhancing the agent's overall capability. This indicates that \texttt{LEASE} has the potential to perform well even with a limited number of preference datasets.}

\subsection{Parameter Analysis}\label{Parameter}

\begin{table}
	\normalsize
	\caption{The results of parameter analysis for the confidence threshold $\kappa_p$ and the uncertainty threshold $\kappa_\tau$.}
	\label{tab_5}
	\centering
	\resizebox{9cm}{!}{
		\begin{tabular}{c|cc|cc}
			\toprule[1pt] 
			\multirow{2}{*}{\textbf{CQL}} & \multicolumn{2}{c|}{$\kappa_{\tau}=0.05$} & \multicolumn{2}{c}{$\kappa_{\tau}=0.08$ } \\
			& $\kappa_{p}=0.85$ &$\kappa_{p}=0.95$& $\kappa_{p}=0.85$ &$\kappa_{p}=0.95$\\
            \midrule
		walker2d-m& $\mathbf{78.4\pm0.9}$& $70.6\pm1.8$ &$72.4\pm1.1$ &$70.3\pm2.1$\\
            pen-expert&$127.0\pm3.4$ &$\mathbf{130.6\pm1.8}$ & $122.9\pm1.8$&$125.8\pm2.4$\\
            \midrule
            \midrule
            \multirow{2}{*}{\textbf{IQL}} & \multicolumn{2}{c|}{$\kappa_{\tau}=0.05$ } & \multicolumn{2}{c}{$\kappa_{\tau}=0.08$} \\
			& $\kappa_{p}=0.85$ &$\kappa_{p}=0.95$& $\kappa_{p}=0.85$ &$\kappa_{p}=0.95$\\
            \midrule
		walker2d-m&$72.6\pm0.1$ & $69.1\pm1.5$& $\mathbf{73.0\pm2.1}$&$68.9\pm3.3$\\
            pen-expert&$107.8\pm4.7$ &$105.4\pm2.4$ &$111.8\pm3.4$ &$\mathbf{113.8\pm6.3}$\\
			\bottomrule[1pt]
		\end{tabular}
	}
\end{table}

The selection mechanism $f(\sigma_0,\sigma_1)$ for the reward model has two critical parameters: the confidence threshold $\kappa_p$ and the uncertainty threshold $\kappa_\tau$. The selection of two parameters depends on multiple factors, including the complexity of the task's reward function, the performance of the pre-trained reward model, and the specific offline RL algorithm employed (since the collected preference data is inherently linked to the policy learned by the algorithm). Given the substantial computational cost inherent in RL training, exhaustive parameter tuning is often impractical. To mitigate this, our work systematically analyzes the impact of key parameters on task performance across multiple algorithms, offering practical guidance for their configuration.

Table \ref{tab_5} presents a comparative analysis of the performance of different offline RL algorithms (CQL and IQL) under varying parameter configurations for both locomotion and manipulation tasks. For the uncertainty threshold $\kappa_\tau$, the results demonstrate that CQL exhibits a smaller uncertainty threshold compared to IQL. This phenomenon can be attributed to CQL's inherent conservatism, which leads to a narrower data distribution when the learned policy interacts with the environmental model (\textit{i.e.}, the generated data is more concentrated within a specific region). Consequently, the reward model exhibits lower prediction uncertainty, allowing for the selection of a reduced uncertainty threshold. For the confidence threshold $\kappa_p$, manipulation tasks require a higher confidence threshold than locomotion tasks. This discrepancy arises from the simpler reward structure of manipulation tasks, which typically rely on discrete variables (See Fig. A.1 in Appendix D). As a result, the pre-trained reward model achieves higher accuracy in manipulation tasks, allowing for the use of a more stringent confidence threshold to enhance the quality of generated data. For additional experiments of the \texttt{LEASE}—such as evaluations of the reward model on alternative tasks, its application within model-based RL algorithms, and an extended comparative analysis—please refer to Appendix D.

	\section{Discussion} \label{discussion}
\subsection{Discussion for Offline PbRL Theory}\label{discussion_1}
In the offline RL field, high sample efficiency refers to the agent's ability to achieve comparable performance under fewer data compared with the performance under large data. In this paper, the data refers to the preference dataset. The labeled preference dataset, each trajectory of length $L$, is collected through real-time human feedback under policy $\mu$, which demands tremendous human effort; thus, the collected cost of preference data is higher than fixed offline data. The unlabeled dataset is generated through a trained transition without real-time interaction under the learned policy $\pi^t$ at time $t$. 

Through data augmentation, the sample efficiency can be significantly reduced. The performance gap caused by reward can be reduced to $\varpi$ under a smaller labeled dataset. Zhan \e \cite{zhan2024provable} also developed a systematic theory for offline PbRL, but did not focus on the generalization performance of the reward model. Moreover, they also introduced the concentrability coefficient $\mathscr{C}_{r}(\pi)$ for PbRL, defined as
\begin{equation*}
\label{yuan_c}
    \max\Bigg\{0,\sup_{r}\frac{\mathbb{E}_{\sigma^0\sim\pi,\sigma^1 \sim \pi_{\text{ref}}}\big[r^*(\sigma^0,\sigma^1)-{r}(\sigma^0,\sigma^1)\big]}{\sqrt{\mathbb{E}_{\sigma^0\sim \mu_0,\sigma^1 \sim \mu_1}\big|r^*(\sigma^0,\sigma^1)-{r}(\sigma^0,\sigma^1)\big|^2}}\Bigg\},
\end{equation*}
where $r(\sigma^0,\sigma^1)=\sum_i [R(s_i^0,a_i^0)-R(s_i^1,a_i^1)]$, $\pi_{\text{ref}}$ is an arbitrary trajectory distribution (usually set as $\mu_1$), and $\mu_0,\mu_1$ are behavior trajectory distribution. However, it is based on trajectory and is difficult to combine with other offline RL theories. The proposed concentrability coefficient in our paper, defined in Eq. (\ref{eq_14}), is based on state-action pairs.

Moreover, the performance gap of offline PbRL between the behavior policy and learned policy is influenced by the offline algorithm itself and the performance of the learned reward (preference) model. However, \cite{zhan2024provable} fails to consider the gap caused by the offline algorithm itself and the sampling error. The theory developed in our paper can be easily combined with another offline algorithm. Then, the method in \cite{zhan2024provable} can learn $\varpi$-optimal policy with a sample complexity of
\begin{equation}
\label{yuan_n}
N=\widetilde{\mathcal{O}}\Bigg(\frac{c^2k^2\mathscr{C}_{r}^2(\widehat{\pi})}{\varpi^2}\log\left(\frac{\mathcal{N}_{r}(1/N)}{\delta}\right)\Bigg),
\end{equation}
where $c>0$ is a universal constant, $k=(\inf_{x\in [-r_{max},r_{max}]}\\ \Phi' (x))^{-1}$, and $\Phi(x)$ is a monotonically increasing link function. Compared with Eq. (\ref{yuan_n}), the sample complexity of \texttt{LEASE} defined in Eq. \eqref{eq_16} contains more useful information. For example, the sample complexity is related to the length $L$ of preference data and the range $R_{max}$ of the reward function, and can be reduced when the learned policy is close to the behavior policy $\mu$.  

\subsection{Discussion for Broader Impacts}

\ttx{\textbf{Actual application.}} \ttx{The high-sample-efficiency preference-based offline RL algorithm proposed in this work aims to learn robot control policies from offline data with minimal human feedback, under the condition of an unknown reward function. This approach is particularly suitable for scenarios where the reward function is difficult to specify mathematically and where human-robot interaction entails high cost. For instance, in the field of exoskeleton robot assistance, objectives such as maximizing assistive comfort and minimizing metabolic cost are challenging to define precisely, and prolonged human-in-the-loop interaction can lead to user fatigue. Therefore, it is imperative to learn a reward model from a limited preference dataset and utilize it to efficiently guide policy optimization.}

\ttx{The deployment of the algorithm for practical control involves three main stages: 1) collecting an offline dataset (containing only $(s, a, s')$ tuples) through methods such as teleoperation, to train a transition model; 2) acquiring a limited set of preference data by having users select their preference between pairs of trajectories, which is then used to learn a reward function; and 3) training the robot control policy offline using the collected dataset and the learned reward model, followed by its deployment to the actual control system. The ultimate goal is to optimize a control policy offline with minimal human-robot interaction efficiently.}

\textbf{Theoretical study.} The theory of \texttt{LEASE} shows that the performance gap between the behavior policy $\mu$ and $\widehat{\pi}$ learned by \texttt{LEASE} includes two part: the gap $\xi$ caused by offline algorithm itself and the gap $\xi_1$ caused by reward model gap (the details see Theorem \ref{the_2}), where the term $\xi_1$ only depends on preference dataset. Moreover, the theory of \texttt{LEASE} is based on state-action pairs, which is consistent with most offline algorithms theory. \ttx{The performance gap in offline PbRL is determined by two independent components, with the first term corresponding to the inherent gap of offline RL itself. Given this gap, the preference-based gap can be directly derived. Therefore, the theory developed in this paper can be easily combined with other offline algorithm theories and used to build a theory of policy improvement guarantee, providing a theoretical basis for offline PbRL and facilitating the development of offline PbRL theory.}

\section{Conclusion}\label{conclusion}
This paper proposes a novel offline PbRL algorithm (\texttt{LEASE}) with high sample efficiency. \texttt{LEASE} can achieve comparable performance under fewer preference datasets. By selecting high-confidence and low-variance data, the stability and accuracy of the reward model are guaranteed. Moreover, this paper provides the theoretical analysis for \texttt{LEASE}, including the generalization of the reward model and policy improvement guarantee. The theoretical and experimental results demonstrate that the data selection mechanism $f(\sigma_0,\sigma_1)$ can effectively improve the performance of the reward model, and the performance learned by \texttt{LEASE} can be guaranteed under fewer preference datasets. \ttx{Future works can focus on the following two aspects: 1) Since the trained transition model may have errors, which will affect the performance of the reward model to some extent, the quality of generated data can be considered in the future to further improve the accuracy of the reward function; 2) There is still a gap between the trained reward model and the real model. Other efficient methods of data augmentation can be used in the future to further narrow this gap.}

        \bibliographystyle{plain}

    \clearpage
\begin{appendix}\label{appendix}

\setcounter{equation}{0}
\renewcommand{\theequation}{A.\arabic{equation}}
\setcounter{figure}{0}
\renewcommand{\thefigure}{A.\arabic{figure}}
\setcounter{table}{0}
\renewcommand{\thetable}{A.\arabic{table}}

\section{Related Proofs}
\subsection{Proof of Proposition \ref{pro_1}} \label{pro_pro_1}

\textbf{{The Proof for Proposition \ref{pro_1}}}: This proof follows the previous work  \cite{xie2024class}. Obviously, the largest pseudo-labeling error $\eta$ in Assumption \ref{ass_2} includes exactly two types of pseudo-labeling error:
\begin{equation}
    \label{ap_1}
    \begin{aligned}
         &\eta_1= \frac{1}{{N}_u}\sum_{j=1}^{{N}_u} \mathbbm{1}\Big[P_ {\psi}\left(\sigma_1^j\succ\sigma_0^j\right)>0.5, y^j=0\Big], \\
         &\eta_2= \frac{1}{{N}_u}\sum_{j=1}^{{N}_u} \mathbbm{1}\Big[P_ {\psi}\left(\sigma_0^j\succ\sigma_1^j\right)>0.5, y^j=1\Big],
    \end{aligned}
\end{equation}
where $\eta_1$ represents the error ratio of classifying the first category as the second category, and $\eta_2$ is the error ratio of classifying the second category as the first category. Then, we prove the gap between the unlabeled loss with pseudo-label $\widehat{\mathcal{L}}_u(\psi)$ and that with true label $\widehat{\mathcal{L'}}_u(\psi)$ from two sides: $\widehat{\mathcal{L}}_u(\psi)\leq \widehat{\mathcal{L'}}_u(\psi)+\eta\Omega$ and $\widehat{\mathcal{L}}_u(\psi)\geq \widehat{\mathcal{L'}}_u(\psi)-\eta\Omega$. Notably, we ignore the selection mechanism in the proof below since it doesn't influence the proof result.

\textit{\textbf{Step 1:}} Proving upper bound : $\widehat{\mathcal{L}}_u(\psi)\leq \widehat{\mathcal{L'}}_u(\psi)+\eta\Omega$.
\begin{equation*}
\label{ap_2}
\begin{aligned}
    &\widehat{\mathcal{L}}_u(\psi)\\
    =&\frac{1}{{N}_u}\sum_{j=1}^{{N}_u}
L\left((\sigma_0^u,\sigma_1^u)^{(j)},\widehat{y}^j\right)\\
=&-\frac{1}{{N}_u}\sum_{j=1}^{{N}_u}
\mathbbm{1}\big[P_ {\psi}\left(\sigma_0^j\succ\sigma_1^j\right)>0.5\big]\log P_ {\psi}\big(\sigma_0^j\succ\sigma_1^j\big)\\
& +\mathbbm{1}\big[P_ {\psi}\left(\sigma_1^j\succ\sigma_0^j\right)>0.5\big]\log P_ {\psi}\big(\sigma_1^j\succ\sigma_0^j\big)\\
\leq&-\frac{1}{{N}_u}\sum_{j=1}^{{N}_u}
\mathbbm{1}\big[y^j=0\big]\log P_ {\psi}\big(\sigma_0^j\succ\sigma_1^j\big)\\
&+\mathbbm{1}\big[y^j=1\big]\log P_ {\psi}\big(\sigma_1^j\succ\sigma_0^j\big)\\
& +\mathbbm{1}\Big[P_ {\psi}\left(\sigma_1^j\succ\sigma_0^j\right)>0.5, y^j=0\Big]\log P_ {\psi}\big(\sigma_0^j\succ\sigma_1^j\big)\\
& + \mathbbm{1}\Big[P_ {\psi}\left(\sigma_0^j\succ\sigma_1^j\right)>0.5, y^j=1\Big]\log P_ {\psi}\big(\sigma_1^j\succ\sigma_0^j\big)\\
\leq& \widehat{\mathcal{L'}}_u(\psi)+\eta_1\max_{j} \Big\{-\log P_ {\psi}\big(\sigma_0^j\succ\sigma_1^j \big)\Big\}\\
&+\eta_2\max_{j} \Big\{-\log P_ {\psi}\big(\sigma_1^j\succ\sigma_0^j \big)\Big\}\\
\leq &\widehat{\mathcal{L'}}_u(\psi)+\eta \Omega.
\end{aligned}
\end{equation*}

\textit{\textbf{Step 2:}} Proving low bound : $\widehat{\mathcal{L}}_u(\psi)\geq \widehat{\mathcal{L'}}_u(\psi)-\eta\Omega$.

\begin{equation*}
\label{ap_3}
\begin{aligned}
    &\widehat{\mathcal{L}}_u(\psi)\\
    =&\frac{1}{{N}_u}\sum_{j=1}^{{N}_u}
L\left((\sigma_0^u,\sigma_1^u)^{(j)},\widehat{y}^j\right)\\
\geq&-\frac{1}{{N}_u}\sum_{j=1}^{{N}_u}
\mathbbm{1}\big[y^j=0\big]\log P_ {\psi}\big(\sigma_0^j\succ\sigma_1^j\big)\\
&+\mathbbm{1}\big[y^j=1\big]\log P_ {\psi}\big(\sigma_1^j\succ\sigma_0^j\big)\\
& -\mathbbm{1}\Big[P_ {\psi}\left(\sigma_1^j\succ\sigma_0^j\right)>0.5, y^j=0\Big]\log P_ {\psi}\big(\sigma_0^j\succ\sigma_1^j\big)\\
& - \mathbbm{1}\Big[P_ {\psi}\left(\sigma_0^j\succ\sigma_1^j\right)>0.5, y^j=1\Big]\log P_ {\psi}\big(\sigma_1^j\succ\sigma_0^j\big)\\
\end{aligned}
\end{equation*}

\begin{equation*}
\begin{aligned}
\geq& \widehat{\mathcal{L'}}_u(\psi)-\eta \Omega.
\end{aligned}
\end{equation*}
Combining Step 1 and Step 2, we can obtain the following result:
\begin{equation}\label{ap_4}
    \Big|\widehat{\mathcal{L}}_u(\psi)-\widehat{\mathcal{L'}}_u(\psi)\Big|\leq \eta\Omega.
\end{equation}
This completes the proof of Theorem \ref{pro_1}.

\subsection{Proof of Theorem \ref{the_1}} \label{pro_the_1}
\begin{lemma}
\label{lem_1}
    Let $\mathcal{F}$ be a family of functions mapping from $\mathcal{X} $ to $\mathbb{R}$ and $\widehat{\mathcal{D}}$ be empirical datasets sampled from an i.i.d. sample ${\mathcal{D}}$ of size $N$. Then, for any $\delta>0$, with probability at least $1-\delta$, the following holds for all $f\in \mathcal{F}$, 
    \begin{equation}
    \label{ap_5}
\Big|\mathbb{E}_{\mathcal{D}}\left[f\right]-\mathbb{E}_{\widehat{\mathcal{D}}}\left[f\right]\Big|\leq 2\widehat{\mathfrak{R}}_{{\widehat{\mathcal{D}}}}(\mathcal{F})+3\sqrt{\frac{\log (2 / \delta)}{2N}},
    \end{equation}
    where $\mathbb{E}_{\widehat{\mathcal{D}}}\left[f\right]=\sum_{i=1}^{m}\left[g(x_i)\right]/m$ is the empirical form of $\mathbb{E}_{\mathcal{D}}\left[f\right]$. This proof can be found in Theorem 3.3 of the work \cite{mohri2018foundations}.
\end{lemma}

\begin{proposition}
    \label{pro_2}
    Let $\mathcal{F}$ be a family of loss functions defined in Eq. (\ref{eq_6}) and $\Pi(\mathcal{R})$ be a family of functions defined in Definition \ref{def_2}. Then, for any sample $\widehat{\mathcal{D}}=\{(\sigma_0,\sigma_1,y)^{(i)}\}_{i=1}^{N}$, the following relation holds between the empirical Rademacher complexities of 
 $\Pi(\mathcal{R})$ and $\mathcal{F}$:
    \begin{equation}
        \label{ap_6}
    \widehat{\Re}_{\widehat{\mathcal{D}}}(\mathcal{F})\leq 2\widehat{\Re}_{\widehat{\mathcal{D}}}\big(\Pi(\mathcal{R})\big).
    \end{equation}
\end{proposition}

\textit{\textbf{Proof.}} According to Definition \ref{def_1}, the empirical Rademacher complexity of $\mathcal{F}$ can be written as:
\begin{equation}
\begin{aligned}
    \label{ap_7}
		&\widehat{\Re}_{\widehat{\mathcal{D}}}(\mathcal{F})\\
  =& \mathbb{E}_{\bb{\sigma}} \bigg[\sup _{R \in \mathcal{R}} \frac{1}{N}\sum_{i=1}^N \sigma_i \big[-
(1-y)\log P\left(\sigma_0\succ\sigma_1\right)\\
&-y\log P\left(\sigma_1\succ\sigma_0\right)\big]\bigg]\\
  \leq& \mathbb{E}_{\bb{\sigma}} \bigg[\sup _{R \in \mathcal{R}} \frac{1}{N}\sum_{i=1}^N \sigma_i \big[
-\log P\left(\sigma_0\succ\sigma_1\right)\big]\bigg]\\
&+\mathbb{E}_{\bb{\sigma}} \bigg[\sup _{R \in \mathcal{R}} \frac{1}{N}\sum_{i=1}^N \sigma_i \big[-\log P\left(\sigma_1\succ\sigma_0\right)\big]\bigg]\\
=&2\widehat{\Re}_{\widehat{\mathcal{D}}}\big(\Pi(\mathcal{R})\big).
\end{aligned}
\end{equation}

This completes the proof of Proposition \ref{pro_2}. 

\textbf{{The Proof for Theorem \ref{the_1}}}: This proof is based on Proposition \ref{pro_1} and \ref{pro_2}. \ttx{Combining Eqs. (\ref{ap_5}) and (\ref{ap_6})}, we can derive that for labeled dataset $\widehat{\mathcal{D}}=\{(\sigma_0,\sigma_1,y)^{(i)}\}_{i=1}^{N}$, the generalization error bound between expected error $\mathcal{L}(\psi)$ and empirical error $\widehat{\mathcal{L}}(\psi)$ holds: 
\begin{equation}
\label{ap_8}
    \mathcal{L}(\psi)\leq \widehat{\mathcal{L}}(\psi)+4\widehat{\mathfrak{R}}_{{\widehat{\mathcal{D}}}}\big(\Pi(\mathcal{R})\big)+3\sqrt{\frac{\log (2 / \delta)}{2N}}.
\end{equation}
However, the dataset for training reward model includes two parts: the labeled dataset $\mathcal{D}_l=\{(\sigma_0^l,\sigma_1^l,y)^{(i)}\}_{i=1}^{N_l}$ and unlabeled dataset $\mathcal{D}_u=\{(\sigma_0^u,\sigma_1^u)^{(i)}\}_{i=1}^{N_u}$, and empirical error $\widehat{\mathcal{L}}_{R}(\psi)=\widehat{\mathcal{L}}_{l}(\psi)+\widehat{\mathcal{L}}_{u}(\psi)$. Let $\widehat{\mathcal{L'}}_{u}(\psi)$ be the empirical error under an unlabeled dataset with true label, then
\begin{equation*}
\label{ap_9}
    \mathcal{L}_{R}(\psi)\leq \widehat{\mathcal{L}}_{l}(\psi)+\widehat{\mathcal{L'}}_{u}(\psi)+4\widehat{\mathfrak{R}}_{{\widehat{\mathcal{D}}}}\big(\Pi(\mathcal{R})\big)+3\sqrt{\frac{\log (2 / \delta)}{2(N_l+N_u)}},
\end{equation*}
where $\widehat{\mathcal{D}}$ is the input combination of labeled and unlabeled dataset, denoted as $\{(\sigma_0,\sigma_1)^{(i)}\}_{i=1}^{N_l+N_u}$. According to Proposition \ref{pro_1}, that is $\widehat{\mathcal{L'}}_u(\psi)\leq \widehat{\mathcal{L}}_u(\psi)+\eta\Omega.$, we can derive
\begin{equation*}
\label{ap_10}
    \mathcal{L}_{R}(\psi)\leq \widehat{\mathcal{L}}_{R}(\psi)+\eta\Omega+4\widehat{\mathfrak{R}}_{{\widehat{\mathcal{D}}}}\big(\Pi(\mathcal{R})\big)+3\sqrt{\frac{\log (2 / \delta)}{2(N_l+N_u)}}.
\end{equation*}
This completes the proof of Theorem \ref{the_1}.

\subsection{Proof of Theorem \ref{the_2}} \label{pro_the_2}

\begin{proposition}
    \label{pro_3}
    Consider a set of trajectories $\{\sigma_0^i,\sigma_1^i\}_{i=1}^N$, each of length $L$, collected from an offline dataset following the distribution $d_T^\mu(s,a)$. Then, for any $\delta\in(0,1]$, with probability at least $1-\delta$, the following holds for all $\widehat{R}\in \mathcal{R}$
    \begin{equation}
    \begin{aligned}
            &\bigg|\mathbb{E}_{(s,a)\sim d_T^\mu(s,a)}\big[R^*(s,a)-\widehat{R}(s,a)\big]\bigg|\\
            \leq &\sqrt{\frac{4C}{NL^2}\log\left(\frac{\mathcal{N}_{\mathcal{R}}(1/N)}{\delta}\right)}+\tx{\sqrt{\frac{4R_{max}^2\log \big(1/\delta \big)}{NL}}},
    \end{aligned}
    \end{equation}
    where $R^*$ is the true reward model and $\widehat{R}$ is the learned reward model. $\mathcal{N}_{\mathcal{R}}(1/N)$ is the bracketing number defined in Definition \ref{def_3}.
\end{proposition}
The proof of Proposition \ref{pro_3} can be found in Appendix \ref{prof_pro_3}. Note that the above reward model is only trained from the offline preference dataset. Through generating more preference data, the upper bound can be tighter. However, the generated data may be inaccurate, which may improve the reward gap. The selection mechanism can effectively solve this problem.

\textbf{{The Proof for Theorem \ref{the_2}}}: This Theorem aims to give the lower bound of term $J(\widehat{\pi},R^*)-J(\mu,R^*)$. We prove this from the ideal case and the actual case. 

\textit{\textbf{1) Ideal case:}} There is no need to consider distribution shift and the empirical error problem. The policy $\widehat{\pi}$ is directly learned by $\widehat{\pi}=\max_{\pi}J(\pi,\widehat{R})$. Thus, $J(\widehat{\pi},\widehat{R})\geq J(\mu,\widehat{R})$ holds. Then, we can derive
\begin{equation}
\label{ap_11}
\begin{aligned}
    &J(\mu,R^*)-J(\widehat{\pi},R^*)\\
    =&J(\mu,R^*)-J(\mu,\widehat{R})+J(\mu,\widehat{R})-J(\widehat{\pi},R^*)\\
    \leq &J(\mu,R^*)-J(\mu,\widehat{R})+J(\widehat{\pi},\widehat{R})-J(\widehat{\pi},R^*)\\
    \leq &\big|J(\mu,R^*)-J(\mu,\widehat{R})\big|+\big|J(\widehat{\pi},R^*)-J(\widehat{\pi},\widehat{R})\big|.
\end{aligned}
\end{equation}
Since $J(\pi,R):=\mathbb{E}_{(s,a)\sim d_{T}^{\pi}(s,a)}[R(s,a)]/(1-\gamma)$, $|J(\pi,R^*)-J(\pi,\widehat{R})|$ can be written as
\begin{equation*}
\label{ap_12}
        \big|J(\pi,R^*)-J(\pi,\widehat{R})\big|
        = \frac{1}{1-\gamma}\Big|\mathbb{E}_{(s,a) \sim d_{T}^{\pi}}\big[R^*(s,a)-\widehat{R}(s,a)\big]\Big|.
\end{equation*}
Then, according to Definition \ref{def_3}, we have
\begin{equation*}
\label{ap_13}
        \Big|J(\pi,R^*)-J(\pi,\widehat{R})\Big|
    = \frac{\mathscr{C}_{\mathcal{R}}(\pi)}{1-\gamma}\Big|\mathbb{E}_{(s,a) \sim d_{T}^{\mu0}}\big[R^*(s,a)-\widehat{R}(s,a)\big]\Big|.
\end{equation*}
By Proposition \ref{pro_3}, the following inequality holds
\begin{equation}
\label{ap_15}
\begin{aligned}
       &J(\mu,R^*)-J(\widehat{\pi},R^*)\leq \frac{1+\mathscr{C}_{\mathcal{R}}(\widehat{\pi})}{1-\gamma}\\
       &\bigg(\sqrt{\frac{4C}{NL^2}\log\left(\frac{\mathcal{N}_{\mathcal{R}}(1/N)}{\delta}\right)}+\tx{\sqrt{\frac{4R_{max}^2\log \big(1/\delta\big)}{NL}}}\bigg). 
\end{aligned}
\end{equation}

\textit{\textbf{2) Actual case:}} Offline RL suffers from a distribution shift problem, and it is a necessity to consider empirical error. Therefore, many offline RL algorithms incorporate conservatism into policy to overcome distribution shift, that is, learning policy through $\widehat{\pi}=\max_{\pi} J(\pi,\widehat{R})-P(\pi)$. Therefore, $J(\widehat{\pi},\widehat{R})\geq J(\mu,\widehat{R})-\xi$ instead of $J(\widehat{\pi},\widehat{R})\geq J(\mu,\widehat{R})$ in actual implantation. The term $\xi$ depends on the algorithm itself. Here, we take CQL as an example. The policy gap is proven in Theorem 3.6 of \cite{kumar2020conservative}. The details are given as follows.
\begin{equation}
\begin{aligned}
      \xi=&\underbrace{\frac{\gamma C_{T,R}R_{max}}{(1-\gamma)^2}\mathbb{E}_{s\sim d_T^\mu(s)}\sqrt{\frac{|\mathcal{A}|(1+D(s))}{|\mathcal{D}(s)|}}}_{:=\xi_1}\\
      &-\underbrace{\frac{\alpha}{1-\gamma}\mathbb{E}_{s\sim d_T^\mu(s)}\bigg[D(s)\bigg]}_{:=\xi_2},  
\end{aligned}
\end{equation}
where $|\cdot|$ denotes cardinality of a specific set, $D(s)=\sum_a \widehat{\pi}(a|s)(\frac{\widehat{\pi}(a|s)}{\mu(a|s)}-1)$ and $C_{T,R}$ is the empirical coefficient. It consists of two terms: the first term $\xi_1$ captures the decrease in policy performance due to sampling error. The second term $\xi_2$ captures the increase in policy performance in the empirical setting \cite{kumar2020conservative}.
Therefore, when considering distribution shift and empirical error, we can derive
\begin{equation*}
\begin{aligned}
    &J(\mu,R^*)-J(\widehat{\pi},R^*)\\
    \leq &J(\mu,R^*)-J(\mu,\widehat{R})+J(\widehat{\pi},\widehat{R})+\xi-J(\widehat{\pi},R^*)\\
    \leq &\xi+\big|J(\mu,R^*)-J(\mu,\widehat{R})\big|+\big|J(\widehat{\pi},R^*)-J(\widehat{\pi},\widehat{R})\big|.
\end{aligned}
\end{equation*}
Furthermore, according to Eq. (\ref{ap_15}), we have
\begin{equation*}
\begin{aligned}
        &J(\mu,R^*)-J(\widehat{\pi},R^*)\leq \xi+\frac{1+\mathscr{C}_{\mathcal{R}}(\widehat{\pi})}{1-\gamma}\\
        &\bigg(\sqrt{\frac{4C}{NL^2}\log\left(\frac{\mathcal{N}_{\mathcal{R}}(1/N)}{\delta}\right)}+\sqrt{\frac{4R_{max}^2\log \big(1/\delta\big)}{NL}}\bigg).
\end{aligned}
\end{equation*}
This completes the proof for Theorem \ref{the_2}.

\subsection{Proof of Proposition \ref{pro_3}}\label{prof_pro_3}
\begin{lemma}
\label{lem_2}
There exists an absolute constant $C$ such that for any $\delta \in (0,1]$, with probability at least $1-\delta$, the following holds for all $\widehat{R}\in \mathcal{R}$
     \begin{equation*}
     \label{ap_19}
        \sum_{i=1}^N \Big({P_{\widehat{R}}(y^i|\sigma_0^i,\sigma_1^i)}-{P_{R^*}(y^i|\sigma_0^i,\sigma_1^i)}\Big)^2 \leq C\log\left(\frac{\mathcal{N}_{\mathcal{R}}(1/N)}{\delta}\right),
    \end{equation*}
    where $y\in \{0,1\}$, $R^*$ is the true reward model and $\widehat{R}$ is the learned reward model. $P_{R}(0~|~\sigma_0,\sigma_1)$ is the probability that $\sigma_0$ is preferable $\sigma_1$ under reward model $R$. $\mathcal{N}_{\mathcal{R}}(1/N)$ is the bracketing number defined in Definition \ref{def_3}. This proof can be found in Lemma 2 of the previous work \cite{zhan2024provable} and Proposition 14 of \cite{liu2022partially}.
\end{lemma}

\textbf{{The Proof for Proposition \ref{pro_3}}}: We prove this Proposition from two steps. Firstly, we bound the difference between $P_{R^*}(\cdot~|~\sigma_0,\sigma_1)$ and $P_{\widehat{R}}(\cdot~|~\sigma_0,\sigma_1)$. Then, we bound the difference between $R^*(s,a)$ and $\widehat{R}(s,a)$.

\textit{\textbf{Step 1:}} Bound the probability difference $|P_{R^*}(\cdot~|~\sigma_0,\sigma_1)- P_{\widehat{R}}(\cdot~|~\sigma_0,\sigma_1)|$. By Cauchy-Schwarz inequality $\big(\sum_i a_i b_i\big)^2\leq \big(\sum_i a_i^2\big)\big(\sum_i b_i^2\big)$, we set $a_i=P_{\widehat{R}}-P_{R^*}$ and $b_i$ that is chosen from $\{-1,1\}$ and satisfies $a_ib_i>0$. Then, we have
\begin{equation}
    \begin{aligned}
     &\sum_{i=1}^N \Big({P_{\widehat{R}}(y^i|\sigma_0^i,\sigma_1^i)}-{P_{R^*}(y^i|\sigma_0^i,\sigma_1^i)}\Big)^2\\
    \geq&\frac{1}{N}\bigg(\sum_{i=1}^N \Big|P_{\widehat{R}}(y^i|\sigma_0^i,\sigma_1^i)-P_{R^*}(y^i|\sigma_0^i,\sigma_1^i)\Big|\bigg)^2.
    \end{aligned}
\end{equation}
Then, by Lemma \ref{lem_2}, the probability difference can be written as
\begin{equation}
\label{ap_16}
    \begin{aligned}
        &\sum_{i=1}^N \Big|P_{\widehat{R}}(y^i|\sigma_0^i,\sigma_1^i)-P_{R^*}(y^i|\sigma_0^i,\sigma_1^i)\Big| \\
        \leq & \sqrt{CN\log\left(\frac{\mathcal{N}_{\mathcal{R}}(1/N)}{\delta}\right)}.
    \end{aligned}
\end{equation}

\begin{figure*}[t]
	\centering
	\includegraphics[width=0.93\textwidth]{fig/Figure_A3.png}
	\caption{The comparison between prediction value by the learned rewards and their ground truths for different methods under (a) pen-expert and (b) hammer-expert datasets. }.
	\label{figA3}
\end{figure*}

\textit{\textbf{Step 2:}} Bound the reward difference $|R^*(s,a)-\widehat{R}(s,a)|$. Let $f(\sigma)=\sum_i R^*(s^i,a^i)\in[-LR_{max},LR_{max}]$, $g(\sigma)=\sum_i \widehat{R}(s^i,a^i)\in[-LR_{max},LR_{max}]$, and $F(x_1,x_2)=e^{x_1}/(e^{x_1}+e^{x_2})$. Then, according to Eq. (\ref{eq_2}), we can derive
\begin{equation}
\begin{aligned}
      &\Big|P_{\widehat{R}}(y|\sigma_0,\sigma_1)-P_{R^*}(y|\sigma_0,\sigma_1)\Big|\\
      =&\Big|F\big(f(\sigma_0),f(\sigma_1)\big)-F\big(g(\sigma_0),g(\sigma_1)\big)\Big|.  
\end{aligned}
\end{equation}

Notably, the above equation only considers the condition $y=0$ since the result of $y=1$ is similar to $y=0$. Then, according to error propagation, we have
\begin{equation}
\label{ap_18}
\begin{aligned}
        &\Big|F\big(f(\sigma_0),f(\sigma_1)\big)-F\big(g(\sigma_0),g(\sigma_1)\big)\Big|\\
        \approx&\Bigg|\frac{\partial F
        }{\partial x_1}\Bigg|\Big|f(\sigma_0)-g(\sigma_0)\Big|+\Bigg|\frac{\partial F
        }{\partial x_2}\Bigg|\Big|f(\sigma_1)-g(\sigma_1)\Big|\\
        \geq &\frac{e^{x_1+x_2}
        }{\big(e^{x_1}+e^{x_2}\big)^2}\Big|\Big(f(\sigma_0)-g(\sigma_0)\Big)+\Big(f(\sigma_1)-g(\sigma_1)\Big)\Big|\\
        \geq &\frac{1}{4}\bigg|\sum_{l=1}^L \sum_{j=0}^1\Big(R^*(s_j^l,a_j^l)-\widehat{R}(s_j^l,a_j^l)\Big)\bigg|.
\end{aligned}
\end{equation}
Then, combining Eqs. (\ref{ap_16}) and (\ref{ap_18}), we can further derive
\begin{equation}
\label{ap_23}
\begin{aligned}
&\bigg|\sum_{i=1}^N \sum_{l=1}^L \sum_{j=0}^1\Big(R^*(s_j^{i,l},a_j^{i,l})-\widehat{R}(s_j^{i,l},a_j^{i,l})\Big)\bigg|\\
\leq &4\sqrt{CN\log\left(\frac{\mathcal{N}_{\mathcal{R}}(1/N)}{\delta}\right)}.
\end{aligned}
\end{equation}

According to Chernoff-Hoeffding bound \cite{hoeffding1994probability}, for independent random variables $X_1,X_2,...,X_n$, with high probability $1-\delta$, the below equation holds
\begin{equation}\label{new}
    \tx{\mathbb{E}[X]\leq \frac{1}{n}\sum_{i=1}^{n}X_i +(b-a)\sqrt{\frac{\log (1/\delta)}{2n}},}
\end{equation}
where $[a,b]$ is the range of values that each $X_i$ can take. Then, for Eq. \eqref{ap_23}, we set $X_i$ as $R^*(s_j,a_j)-R(s_j,a_j)$ and $X_i \in[-2R_{max},2R_{max}])$. Then, the following equation holds.
\begin{equation}
\begin{aligned}
        &\bigg|\mathbb{E}_{(s,a)\sim d_T^\mu(s,a)}\big[R^*(s,a)-\widehat{R}(s,a)\big]\bigg|\\
        \leq &\sqrt{\frac{4C}{NL^2}\log\left(\frac{\mathcal{N}_{\mathcal{R}}(1/N)}{\delta}\right)}+\tx{\sqrt{\frac{4R_{max}^2\log \big(1/\delta\big)}{NL}}},
\end{aligned}
\end{equation}
where $d_T^\mu(s,a)$ is the distribution of offline dataset. The above equation holds since the trajectories $\{\sigma_0^i,\sigma_1^i\}_{i=1}^N$ are collected from offline dataset. This completes proof of Proposition \ref{pro_3}.

\begin{table*}
\caption{Comparison results between \texttt{URLHF} using fewer preference data and \texttt{LEASE}, where the offline RL algorithm is CQL and \texttt{URLHF} using fewer data is denoted as {$\texttt{URLHF}^*$}.}
 \label{tab_100}
	\renewcommand{\arraystretch}{0.8}
	\centering
	\resizebox{0.7\linewidth}{!}
	{
		\begin{tabular}{c|cccc}
			\toprule[1pt] 
\textbf{Task Name} & \texttt{$\textbf{URLHF}^*$}&\texttt{\textbf{URLHF}}&\texttt{\textbf{FEWER}} & \texttt{\textbf{LEASE}} \\
\midrule
walker2d-m & $76.1\pm0.8$&$76.0\pm0.9$ & $77.4\pm0.6$ & $\mathbf{78.4\pm0.9}$\\
walker2d-m-e& $86.8\pm18.6$& $92.8\pm22.4$ & $77.7\pm0.3$ & $\mathbf{98.6\pm 18.1}$ \\
hopper-m& $\mathbf{56.6\pm2.4}$ & $54.7\pm3.4$ & $55.8\pm2.8$ & $\mathbf{56.5\pm0.6}$\\
hopper-m-e& $55.3\pm0.9$& $\mathbf{57.4\pm4.9}$ & $53.6\pm0.9$ & $56.4\pm0.8$ \\
halfcheetah-m&$43.3\pm0.2$ & $\mathbf{43.4\pm 0.1}$ & $\mathbf{43.5\pm0.1}$ & $\mathbf{43.5\pm0.1}$ \\
halfcheetah-m-e& $58.9\pm2.3$&$\mathbf{62.7\pm7.1}$ & $48.3\pm0.7$ & $53.2\pm3.1$ \\
\midrule
\textbf{Mujoco Average}& $62.8\pm4.2$& $\mathbf{64.5\pm6.5}$ & $59.4\pm0.9$ & $\mathbf{64.4\pm4.0}$ \\
\midrule
pen-human & $\mathbf{17.7\pm13.0}$ & $9.8\pm14.1$ & $0.5\pm3.0$ & $3.8\pm4.6$ \\
pen-expert& $114.6\pm53.7$&  $\mathbf{138.3\pm5.2}$ & $128.1\pm0.7$ & $130.6\pm1.8$ \\
door-human & $1.7\pm1.1$ & $\mathbf{4.7\pm5.9}$ & $0.2\pm1.0$ & $\mathbf{4.7\pm8.8}$  \\
door-expert& $\mathbf{103.3\pm0.5}$& $\mathbf{103.9\pm0.8}$ & $103.0\pm0.9$& $\mathbf{103.2\pm 0.7}$  \\
hammer-human & $0.7\pm0.1$& $\mathbf{0.9\pm0.3}$ & $0.3\pm0.0$ & $0.3\pm0.0$  \\
hammer-expert& $117.4\pm2.7$& $120.2\pm6.8$ & $124.1\pm2.1$ & $\mathbf{126.3\pm1.2}$ \\
\midrule
\textbf{Adroit Average}& $59.2\pm11.8$& $\mathbf{63.0\pm5.5}$ & $59.4\pm1.3$ & $61.5\pm2.9$\\
\bottomrule[1pt]
\end{tabular}
	}
\end{table*}

\subsection{Other Experimental Results}
These part presents supplementary experimental results, including: 1) the performance evaluation of the reward model on other tasks, 2) a comparative analysis between the proposed method \texttt{LEASE} and \texttt{URLHF} when trained on identical datasets, 3) an investigation of our framework's effectiveness in model-based offline RL settings, and 4) further analysis on the performance on benchmark tasks. The comprehensive results provide additional empirical validation of the performance of the proposed method.

\subsubsection{The other comparison results for reward model performance}
Fig. \ref{figA3} shows the comparison between the prediction value by the learned rewards and their ground truths for \texttt{LEASE}, \texttt{FEWER}, and \texttt{FRESH} under Adroit tasks, where the offline algorithm is IQL. \ttx{For evaluation, we randomly sample $500$ data points from the unlabeled dataset that are not seen during the training phase, as illustrated in Fig. 4. Because rewards in Adroit tasks are limited to a few specific values (\textit{i.e.}, discrete states), while rewards in Mujoco tasks are continuous, Fig. \ref{figA3} is presented as multiple scatter lines, in contrast to Fig. 4.} As evidenced by the experimental results shown in this figure, the reward model trained under the LEASE framework achieves the smallest variance in predicted reward values, which shows that the data augmentation and selection mechanism can improve the reward model's performance.

\subsubsection{The further comparison results between \texttt{LEASE} and \texttt{URLHF}}
Considering that it is crucial to include a comparison with URLHF using the same amount of data as LEASE. To further show the superior performance of the proposed method, we compare \texttt{LEASE} to the baseline algorithm \texttt{URLHF} with the same amount of data as \texttt{LEASE}, where the latter method is denoted as $\text{URLHF}^*$.  Table \ref{tab_100} shows that the average performance of \texttt{LEASE} is superior to that of the baseline algorithm \texttt{URLHF} using the same amount of data with \texttt{LEASE}. The experimental results demonstrate that the proposed algorithm \texttt{LEASE} achieves a performance improvement of approximately 3.9\% compared to $\text{URLHF}^*$. This further validates the effectiveness and advantages of the proposed framework and method.

\begin{table}
\caption{ Comparison results of offline RL algorithms under the designed framework.}
 \label{tab_model}
	\renewcommand{\arraystretch}{0.8}
	\centering
	\resizebox{0.99\linewidth}{!}
	{
		\begin{tabular}{c|ccc}
			\toprule[1pt] 
\textbf{Task Name} & $\textbf{CQL}^*$&$\textbf{IQL}^*$&$\textbf{COMBO}^*$ \\
\midrule
walker2d-m &$78.4\pm 0.9$ & $73.0\pm2.1$&$71.6\pm2.4$ \\
walker2d-m-e&$98.6\pm 18.1$ & $108.1\pm0.5$&$79.1\pm1.1$ \\
hopper-m& $56.5\pm0.6$ &$56.0\pm0.5$ & $54.8\pm0.9$\\
hopper-m-e& $56.4\pm0.8$& $55.9\pm1.9$&$54.9\pm1.1$ \\
halfcheetah-m&$43.5\pm0.1$ & $43.0\pm0.3$& $42.9\pm0.1$\\
halfcheetah-m-e&$53.2\pm3.1$ & $60.2\pm1.2$& $73.8\pm7.0$\\
\midrule
\textbf{Mujoco Average}&$64.4\pm 4.0$ &$66.4\pm1.1$ &$62.9\pm2.1$ \\
\bottomrule[1pt]
\end{tabular}
	}
\end{table}

\subsubsection{ The results for \texttt{LEASE} under model-based RL offline algorithm} To enhance the generalizability of the proposed framework, we extended our experimental validation beyond model-free offline RL approaches (such as CQL and IQL) to include model-based offline RL methodologies.
Table \ref{tab_model} shows the results of COMBO \cite{yu2021combo} under the designed framework, where $*$ denotes the results of offline RL algorithms under our designed framework. For COMBO hyperparameters, the probability confidence $\kappa_p$, and uncertainty variance $\kappa_\tau$ are set as $0.85$ and $0.05$ for all tasks, respectively. \ttx{It is important to note that, within our framework, model-based methods do not necessarily outperform model-free approaches. Model-based offline reinforcement learning typically emphasizes learning conservative policies through techniques such as regularizing Q-values or penalizing rewards, which mitigates the impact of inaccuracies in the model-generated data. However, these methods rely heavily on the accuracy of the reward model. Consequently, model-based approaches are more sensitive to errors in the reward model than their model-free counterparts, which may constrain potential improvements in agent performance. Therefore, model-based offline reinforcement learning generally requires a higher degree of reward model accuracy compared to model-free methods.}

\subsubsection{Further analysis for the results on benchmark tasks}
Table \ref{figA3} shows that \texttt{LEASE} doesn't achieve better results on all tasks. Compared to URLHF*, LEASE shows significant improvements on several tasks, such as walker2d-m-e, pen-expert, and hammer-expert. For tasks like walker2d-m, hopper-m-e, halfcheetah-m, and door-human, the performance shows slight improvements. It is widely accepted that the current improved RL algorithms struggle to perform well across all tasks. The network architectures and parameters of URLHF and LEASE differ, which could also lead to the performance discrepancy. It can also be observed that in the hopper, halfcheetah, and door environments, the performance improvement of LEASE over URLHF* is minimal, and in some cases, even lower. This is because the accuracy of the reward function in these environments does not have a significant impact on performance. Table \ref{number11} has validated this conclusion. This table shows the performance comparison of the agent in the above three environments using official URLHF code, with different numbers of preference data $100$, $500$, and $1000$.



\begin{table}
\caption{ Comparison results between different numbers of preference data.}
 \label{number11}
	\renewcommand{\arraystretch}{0.9}
	\centering
	\resizebox{0.98\linewidth}{!}
	{
		\begin{tabular}{c|ccc}
			\toprule[1pt] 
\textbf{Number}&hopper-medium&  halfcheetah-medium& door-expert\\
\midrule
$100$& $56.6\pm2.4$& $43.3\pm0.2$ &$103.3\pm0.5$ \\
$500$& $55.8\pm0.1$& $43.4\pm0.2$ & $103.8\pm1.0$ \\
$1000$& $55.6\pm2.8$& $43.3\pm0.2$ & $103.6\pm0.4$ \\
\bottomrule[1pt]
\end{tabular}
}
\end{table}


\end{appendix}

\end{document}